%% file: main.tex
\documentclass[10pt,twocolumn,letterpaper]{article}

\usepackage[pagenumbers]{iccv} 

\input{preamble}

\definecolor{iccvblue}{rgb}{0.21,0.49,0.74}
\usepackage[pagebackref,breaklinks,colorlinks,allcolors=iccvblue]{hyperref}
\usepackage[accsupp]{axessibility}

\def\paperID{12371} 
\def\confName{ICCV}
\def\confYear{2025}

\title{Semantic Watermarking Reinvented: Enhancing Robustness and Generation Quality with Fourier Integrity}

\author{Sung Ju Lee \quad \quad \quad Nam Ik Cho\\
\\
Dept. of ECE \& INMC, Seoul National University, Korea\\
{\tt\small thomas11809@snu.ac.kr, nicho@snu.ac.kr}
}

\begin{document}
\maketitle
\input{sec/0_abstract}    
\input{sec/1_intro}
\input{sec/2_related}
\input{sec/3_preliminaries}
\input{sec/4_methods}
\input{sec/5_experiments}

\input{sec/6_conclusion}
\clearpage
\section*{Acknowledgements}
This research was supported by the 2024 AI Semiconductor Application/Demonstration Support Program of the Ministry of Science and ICT and the National IT Industry Promotion Agency (NIPA) with Markany Co., Ltd. as the lead organization, and in part by the BK21 FOUR program of the Education and Research Program for Future ICT Pioneers, Seoul National University in 2025.
{
    \small
    \bibliographystyle{ieeenat_fullname}
    \bibliography{main}
}

\input{sec/X_suppl}

\end{document}

%% file: preamble.tex
\usepackage{graphicx}
\usepackage{amsmath}
\usepackage{amssymb}
\usepackage{booktabs}
\usepackage{blindtext}
\usepackage{multirow}
\usepackage{makecell}
\usepackage{colortbl}



%% file: sec/0_abstract.tex
\begin{abstract}
Semantic watermarking techniques for latent diffusion models (LDMs) are robust against regeneration attacks, but often suffer from detection performance degradation due to the loss of frequency integrity. To tackle this problem, we propose a novel embedding method called Hermitian Symmetric Fourier Watermarking (SFW), which maintains frequency integrity by enforcing Hermitian symmetry.
Additionally, we introduce a center-aware embedding strategy that reduces the vulnerability of semantic watermarking due to cropping attacks by ensuring robust information retention. To validate our approach, we apply these techniques to existing semantic watermarking schemes, enhancing their frequency-domain structures for better robustness and retrieval accuracy.
Extensive experiments demonstrate that our methods achieve state-of-the-art verification and identification performance, surpassing previous approaches across various attack scenarios. Ablation studies confirm the impact of SFW on detection capabilities, the effectiveness of the center-aware embedding against cropping, and how message capacity influences identification accuracy. Notably, our method achieves the highest detection accuracy while maintaining superior image fidelity, as evidenced by FID and CLIP scores.
Conclusively, our proposed SFW is shown to be an effective framework for balancing robustness and image fidelity, addressing the inherent trade-offs in semantic watermarking. 
Code available at \texttt{github.com/thomas11809/SFWMark}.
\end{abstract}

%% file: sec/1_intro.tex
\begin{figure}[t]
\centering
\includegraphics[width=\linewidth]{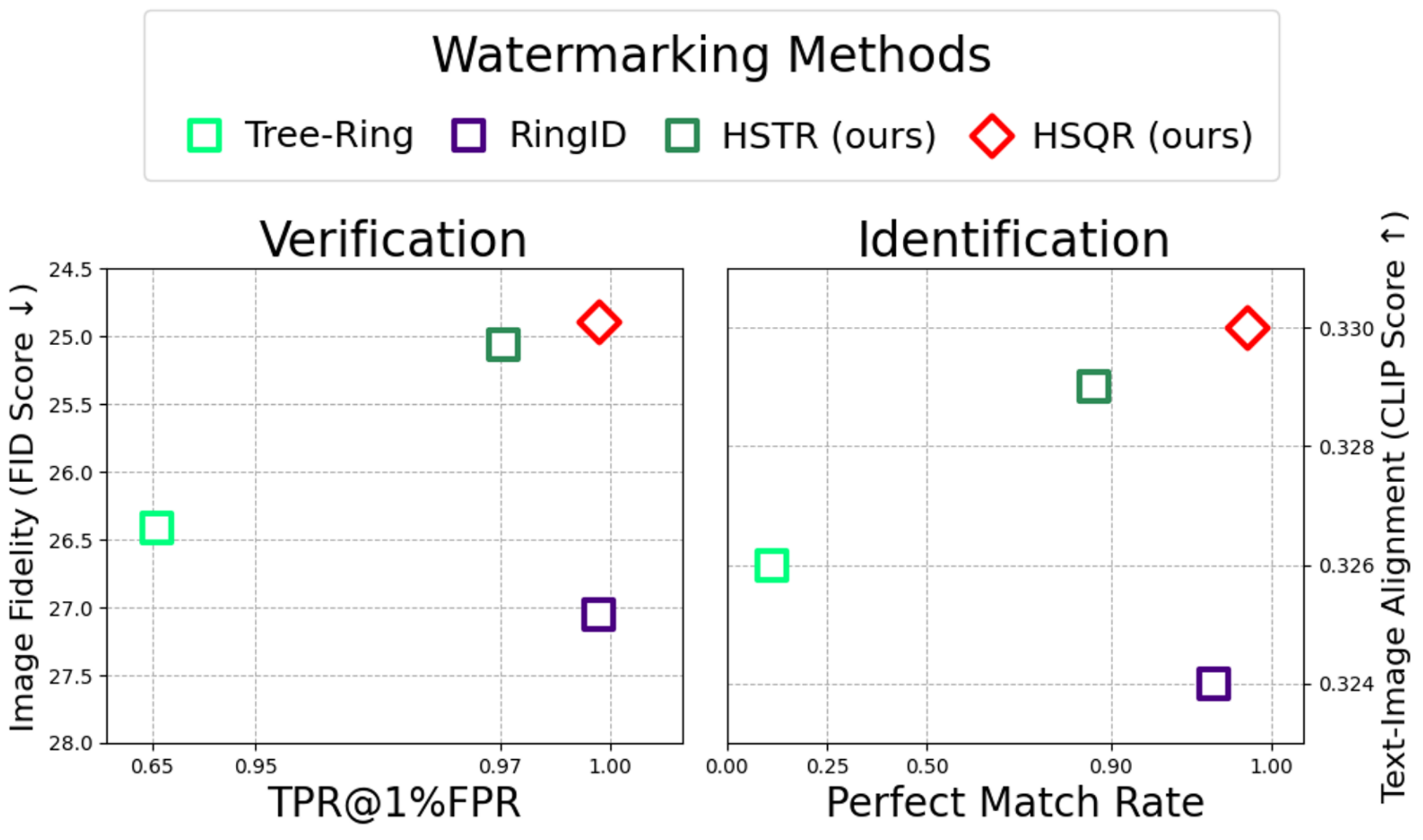}
\caption{Summary of watermarking performance across different semantic watermarking methods, as detailed in \cref{sec:exp-baselines}. All methods follow the \textit{merged-in-generation} scheme with no additional processing time. Verification is evaluated using TPR@1\%FPR (True Positive Rate at 1\% False Positive Rate), while identification is assessed by Identification Accuracy (Perfect Match Rate). The proposed approaches achieve the best balance between detection robustness and image fidelity.}
\label{fig:trade-off}
\end{figure}

\section{Introduction}
\label{sec:intro}
With the advancement of diffusion generative models~\cite{ho2020denoising, song2020denoising, dhariwal2021diffusion, nichol2021glide, ramesh2022hierarchical, saharia2022photorealistic}, their generated images are being increasingly used across various fields, including creative works, entertainment, and advertisement. In particular, the open-source release of large-scale language-image (LLI) models like Stable Diffusion~\cite{rombach2022high} has led to an exponential growth in script-based image generation and editing technologies, resulting in a surge of generative content. This has raised new concerns, such as the copyright of AI-generated content and the tracking of images created with malicious intent. As a solution to these problems, the technique of embedding invisible watermarks into content has been proposed.

Digital content watermarking has been extensively studied through both classical signal processing techniques~\cite{van1994digital, tian2003reversible, cox1997secure, barni1998dct, barni2001improved, ruanaidh1998rotation, wolfgang1999perceptual, cox2007digital, dejey2010improved, li2015dither} and deep learning-based approaches~\cite{zhu2018hidden,zhang2019robust, ahmadi2020redmark, lee2020convolutional, fang2020deep, tancik2020stegastamp, luo2020distortion, zhao2022dari, fang2022end, fernandez2022watermarking}.
Recently, watermarking techniques that directly intervene in the generation process of diffusion models~\cite{fernandez2023stable, bui2023rosteals, zhang2024editguard, asnani2024promark, meng2024latent, rezaei2024lawa, zhang2024training, ci2024wmadapter, feng2024aqualora, lei2024diffusetrace} have been explored; however, these approaches require additional setup or configuration, adding more complexity.

Several studies have explored embedding watermarks directly into the latent representation, eliminating the need for external models. These include a steganographic approach that constructs latent noise with stream cipher-randomized bits~\cite{yang2024gaussian} and semantic watermarking methods that embed geometric patterns in the Fourier frequency domain of the latent representation~\cite{wen2024tree, ci2024ringid, zhang2025attack}.
Meanwhile, research on pixel-level perturbation-based watermarking has highlighted its vulnerability to regeneration attacks~\cite{zhao2023generative, zhao2025invisible}, demonstrating that semantic watermarking serves as a more robust alternative against such attacks. However, the aforementioned semantic watermarking methods in the latent Fourier domain, which are the focus of this paper, suffer from degraded detection accuracy and generative quality due to their lack of frequency integrity preservation.

In this context, we propose a novel semantic watermarking framework that enhances both detection robustness and image quality by preserving frequency integrity in the latent Fourier domain. Our method is named Hermitian Symmetric Fourier Watermarking (SFW), which ensures that watermark embeddings maintain statistical consistency with the latent noise distribution, leading to improved retrievability and stability in generative models.
Additionally, we incorporate a center-aware embedding strategy that enhances robustness against cropping attacks by embedding watermarks in a spatially resilient region of the latent representation.
We comprehensively evaluate our method across various attack scenarios, including signal processing distortions, regeneration attacks, and cropping attacks. 
Experimental results, as illustrated in \cref{fig:trade-off}, demonstrate that our method achieves state-of-the-art detection accuracy in both verification and identification tasks while simultaneously maintaining superior generative quality, as evidenced by FID and CLIP score evaluations.
Our contributions can be summarized as follows:
\begin{itemize}
    \item We propose Hermitian SFW to ensure frequency integrity, leading to improved watermark detection and generative quality.
    \item We introduce center-aware embedding, which significantly enhances robustness against cropping attacks.
    \item We present extensive evaluations demonstrating that our approach outperforms existing baselines in detection accuracy and generative quality.
\end{itemize}

%% file: sec/2_related.tex
\section{Related Works}
\noindent
\textbf{Digital Watermarking.} 
Digital watermarking aims to achieve a balance between high embedding capacity and visual quality when inserting invisible watermarks while also ensuring robustness against various attacks~\cite{wan2022comprehensive}. This field originally began with the adoption of classical signal processing techniques. Embedding the watermark in the spatial domain is a straightforward and intuitive watermark insertion~\cite{van1994digital, tian2003reversible}, but it tends to be less robust against attacks such as filtering or compression. On the other hand, studies utilizing the frequency domain have demonstrated resilience against JPEG compression by embedding watermarks in the low and middle frequency bands~\cite{cox1997secure, barni1998dct, barni2001improved, ruanaidh1998rotation, wolfgang1999perceptual, cox2007digital, dejey2010improved, li2015dither, miyazaki2001analysis, wen2009research, arya2010survey}.

\noindent
\textbf{Watermarks for Latent Diffusion Models.} 
Latent diffusion models (LDMs), such as Stable Diffusion~\cite{rombach2022high}, conduct the diffusion process in a lower-dimensional latent space rather than directly on high-resolution images. Recent research has focused on techniques for integrating watermarks into the latent diffusion process to ensure traceability and robustness. However, many of these approaches rely on model fine-tuning~\cite{fernandez2023stable, asnani2024promark, ci2024wmadapter, feng2024aqualora, lei2024diffusetrace} or require separately trained encoders and decoders~\cite{fernandez2023stable, bui2023rosteals, zhang2024editguard, meng2024latent, rezaei2024lawa, zhang2024training, ci2024wmadapter, feng2024aqualora, lei2024diffusetrace, guo2024freqmark, cui2025diffusionshield} to facilitate watermark embedding and detection. These dependencies constrain the flexibility of watermarking, limiting their applicability across diverse models.

\begin{figure}[t]
\centering
\includegraphics[width=\linewidth]{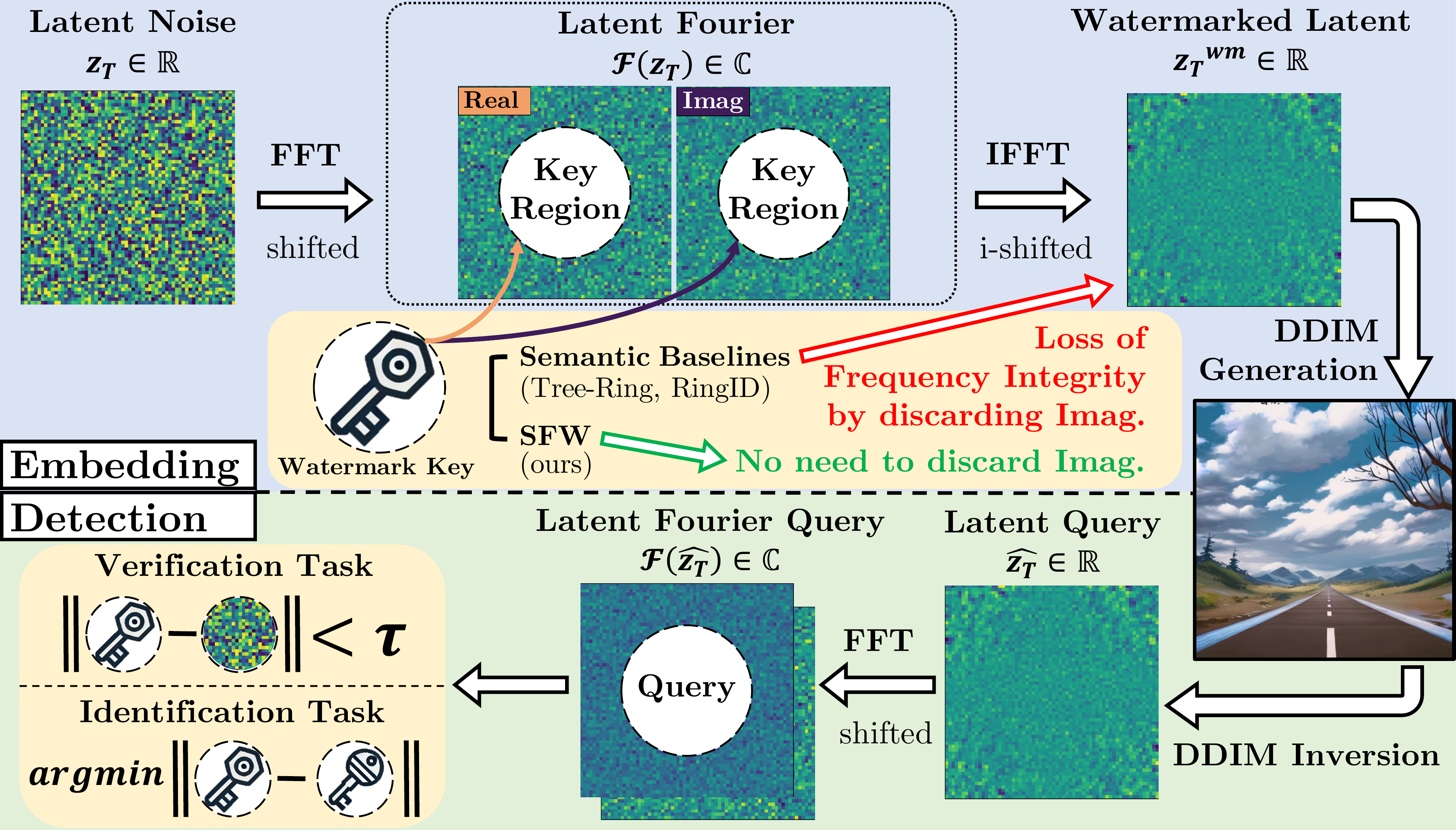}
\caption{Overview of the semantic watermarking process in the latent Fourier domain using the \textit{merged-in-generation} scheme.}
\label{fig:wm-pipeline}
\end{figure}

\noindent
\textbf{Semantic Watermarks in the Latent Fourier Domain.} 
Several studies have explored embedding semantic watermarks in the Fourier domain of latent vectors using a \textit{merged-in-generation} scheme. Wen \etal~\cite{wen2024tree} introduced a tree-ring-shaped watermark constructed with Gaussian-distributed radii, while Ci \etal~\cite{ci2024ringid} proposed a modified pattern with high-energy signed constant rings for watermark identification, along with an additional random noise key embedded in a separate channel.
On the other hand, Zodiac~\cite{zhang2025attack} followed a post-hoc approach, embedding a tree-ring pattern into generated images through multiple iterations of latent vector optimization and diffusion-based synthesis. However, in addition to its high computational cost, this method suffers from low practicality, as it relies on extensive linear interpolation between the original and optimized images to artificially improve visual quality.

In these methods, when performing inverse Fourier transform, the imaginary component in the spatial domain is discarded, leading to a distorted frequency representation. As a result, the real component of the key region retains only partial information, while the imaginary component is almost entirely lost, creating an empty key region in the frequency domain. Since detection relies on analyzing this incomplete key region, the process is inherently limited, resulting in suboptimal retrieval performance.

\cref{fig:wm-pipeline} illustrates the overall pipeline of semantic watermarking with the \textit{merged-in-generation} scheme. The embedding process begins with the Fourier transform applied to the latent noise, generating the latent Fourier representation. A watermark key is then embedded into a designated key region, followed by the inverse Fourier transform, producing the watermarked latent vector. Finally, text-guided image generation synthesizes the watermarked image.
For detection, the process starts with a clean or attacked image, from which the latent query is obtained via DDIM inversion~\cite{song2020denoising}. The presence of a watermark is determined by analyzing the query key region in the latent representation. Further details on detection tasks and evaluation metrics are provided in \cref{sec:task-formulation}.

%% file: sec/3_preliminaries.tex
\begin{figure}[t]
\centering
\begin{subfigure}{0.49\linewidth}
\centering
\includegraphics[width=\linewidth]{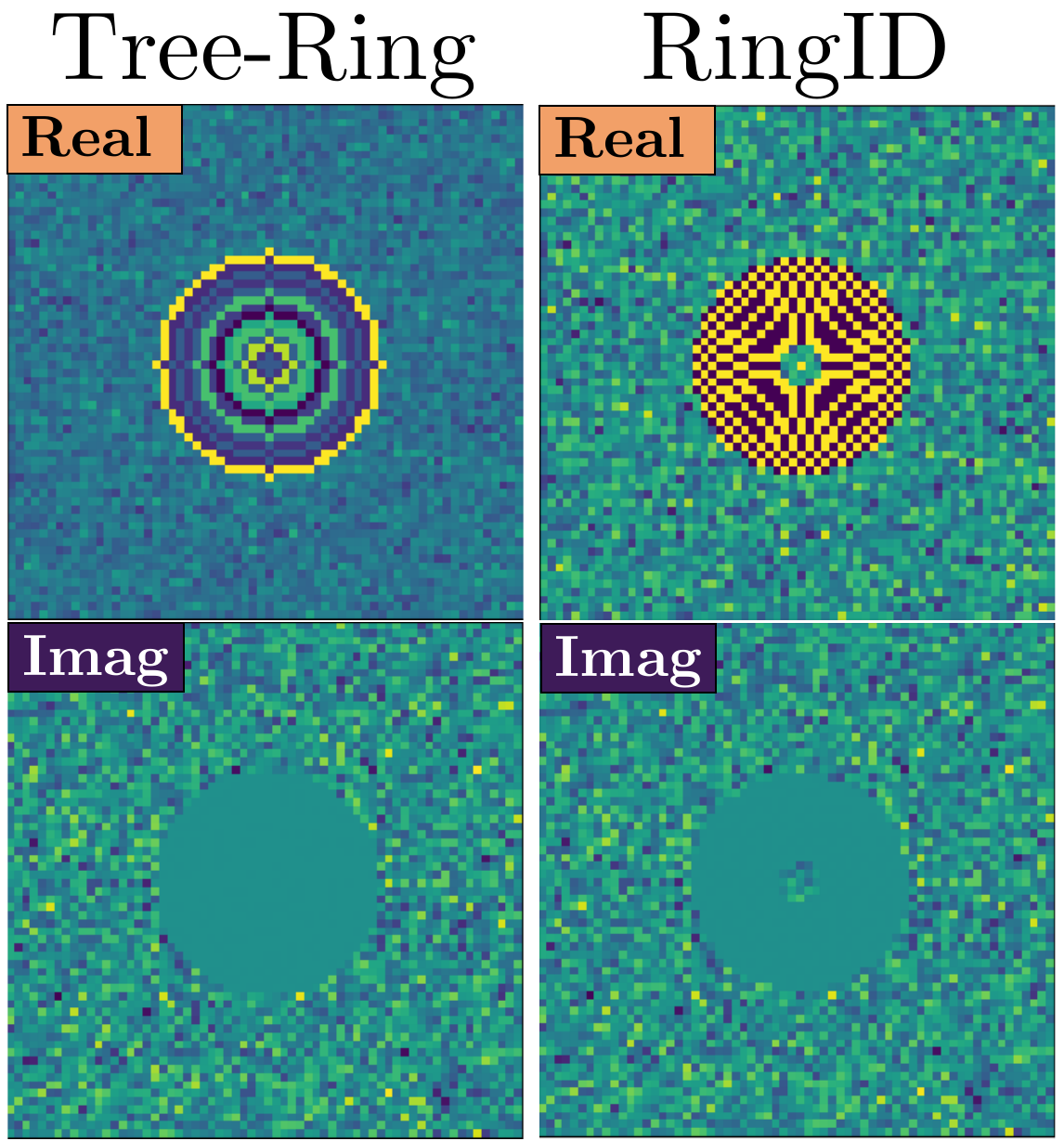}
\caption{Baselines~\cite{wen2024tree, ci2024ringid}.}
\label{fig:wm-fft-a}
\end{subfigure}
\hfill
\begin{subfigure}{0.49\linewidth}
\centering
\includegraphics[width=\linewidth]{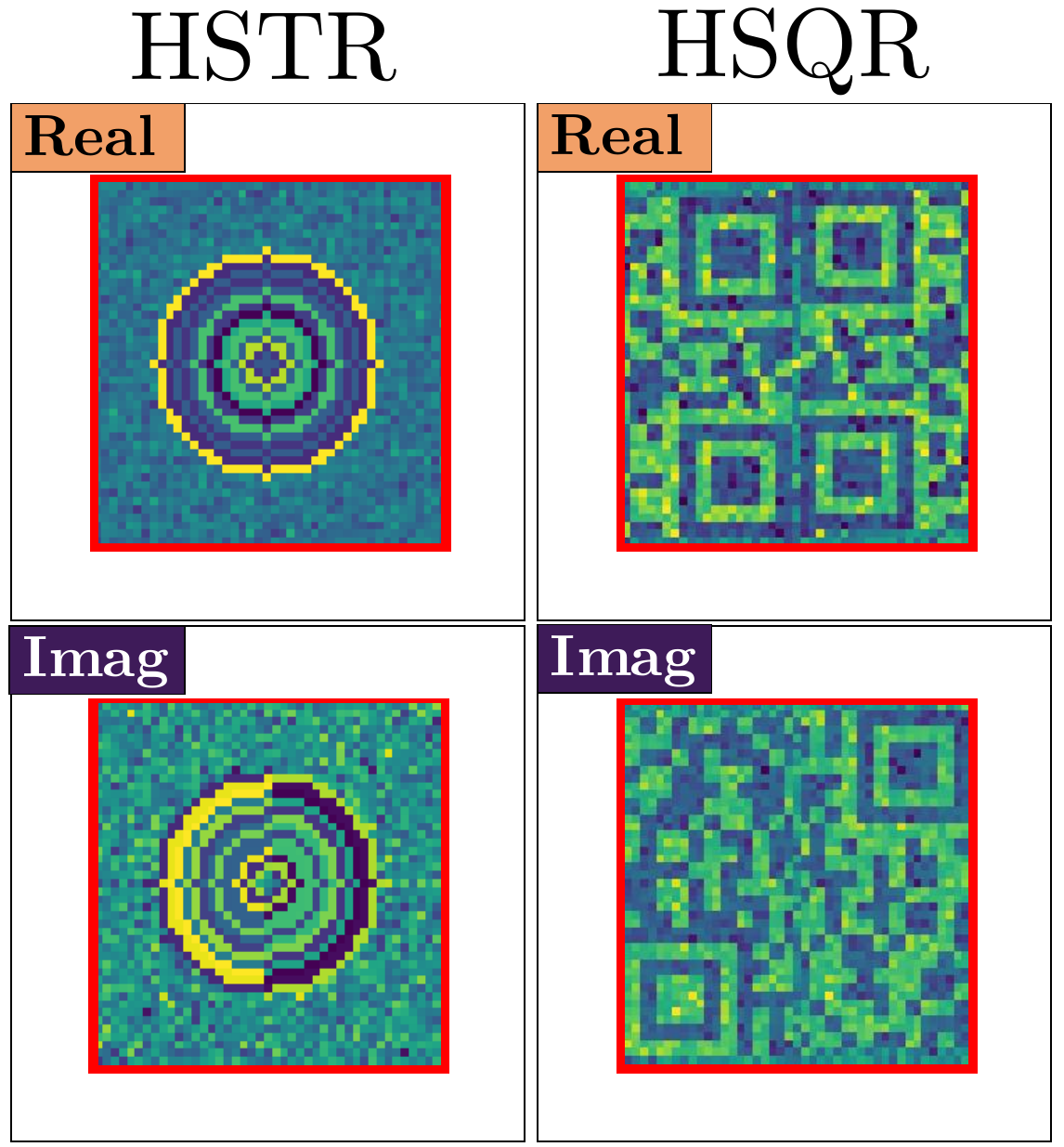}
\caption{Proposed methods.}
\label{fig:wm-fft-b}
\end{subfigure}
\caption{Examples of various semantic watermarking patterns.}
\label{fig:wm-fft}
\end{figure}

\section{Preliminaries}
\subsection{Task Formulation}
\label{sec:task-formulation}
The performance of watermark detection is evaluated following \textit{RingID}'s formulation for watermark verification and identification tasks~\cite{ci2024ringid}. The metric used to measure the distance $d$ is the $L_{1}$ distance, calculated specifically for the key region where the watermark is embedded.

\noindent
\textbf{Verification.}
The objective of verification is to determine whether a watermark is present in an image by analyzing the distance between the reference key and the key region of the query in the latent Fourier domain. Let $\hat{w}$ denote the watermarked latent Fourier key, and $\hat{u}$ denote the unwatermarked latent Fourier null key. Verification is based on comparing the distances between the reference key $w$ and the watermarked/unwatermarked keys, i.e., $d(\hat{w}, w) \neq d(\hat{u}, w)$. Performance is assessed using statistical metrics derived from the ROC curve, considering different distance thresholds.

\noindent
\textbf{Identification.}
In identification, given that a watermark is already embedded, the task is to accurately determine the embedded information. This is achieved by computing the distance between the watermarked key $\hat{w}$ and multiple reference keys $w_{i}$. Performance is evaluated based on the accuracy of the estimated message index, defined as $\hat{i} = \arg\min_{i} d(\hat{w}, w_{i})$.

\begin{figure*}[t]
\centering
\includegraphics[width=\linewidth]{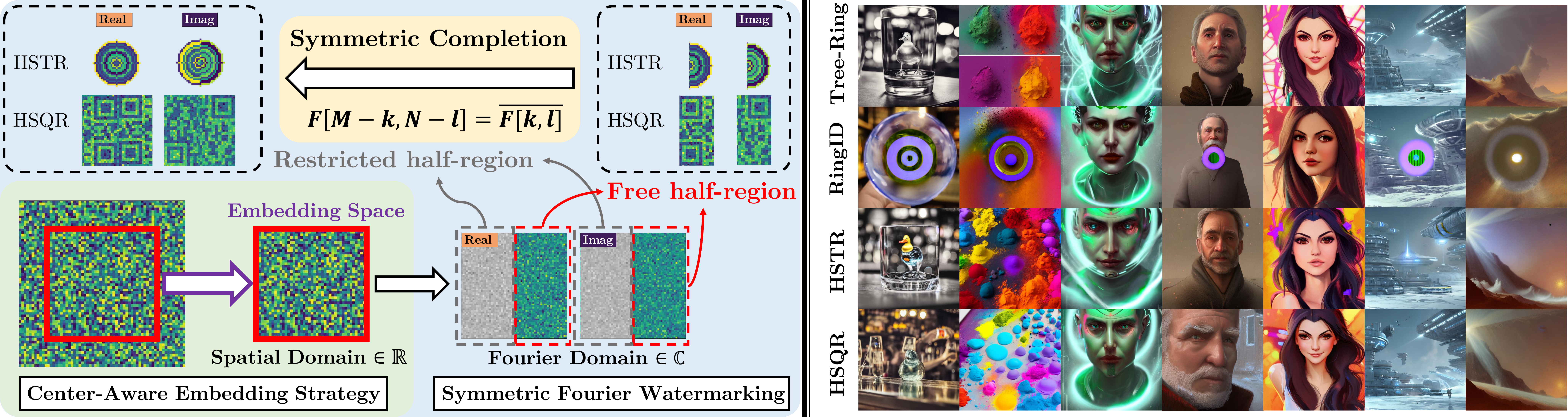}
\caption{Overview of the proposed framework and qualitative results. \textbf{(Left)} The key components of our approach: Symmetric Fourier Watermarking (SFW) (blue region) and Center-Aware Embedding Strategy (green region). \textbf{(Right)} Qualitative results of Tree-Ring, RingID, HSTR, and HSQR. Notably, RingID exhibits visible \textit{ring-like} artifacts, highlighting that its high-energy pattern disrupts generative quality, unlike other \textit{merged-in-generation} semantic watermarking methods that achieve a better balance between robustness and image fidelity.}
\label{fig:concept}
\end{figure*}

\subsection{Fourier Considerations for Real Latent Noise}
This section introduces the mathematical properties of the Fourier domain and highlights important considerations when embedding semantic watermarks.

\noindent
\textbf{Hermitian Symmetry for Real Signals.}
To obtain a real-valued signal after inverse Fourier transform, modifications in the frequency domain must maintain Hermitian symmetry about the DC center:
\begin{equation}
F[M - k, N - l] = \overline{F[k, l]} ,
\end{equation}
where $F$ represents the discrete Fourier transform, $M$ and $N$ denote the dimensions of the signal in the spatial domain along the row and column axes, respectively.
Failure to preserve this symmetry introduces undesired imaginary components in the spatial domain, which are incompatible with the real-valued latent noise required for diffusion models. 

\noindent
\textbf{Impact of Ignoring Hermitian Symmetry.}
In existing baselines~\cite{wen2024tree, ci2024ringid, zhang2025attack}, this issue is simply addressed by discarding the imaginary component, resulting in a loss of frequency integrity. As illustrated in \cref{fig:wm-fft-a}, this disruption affects both real and imaginary components of the original watermark pattern. While the real component deviates from its intended structure due to frequency distortion, the imaginary component suffers even greater degradation, often becoming entirely empty. This significantly limits detection performance by preventing the effective use of frequency information in the imaginary domain. The impact of the frequency loss is further examined in \cref{sec:ablation-scope}.

\noindent
\textbf{Preservation of Gaussianity.}
A real Gaussian noise signal in the spatial domain transforms into a complex Gaussian noise signal in the frequency domain while maintaining its statistical properties:
\begin{equation}
f[m, n] \sim \mathcal{N}(0, \sigma^2) \implies F[k, l] \sim \mathcal{CN}(0, MN \sigma^2),
\end{equation}
where $f$ represents the real-valued Gaussian noise signal in the spatial domain.
When embedding watermarks in the frequency domain, some perturbation is inevitably introduced, altering the original distribution of complex Gaussian noise. However, if Hermitian symmetry is not preserved, the inverse Fourier transform produces incomplete noise characteristics in the spatial domain, as imaginary components must be discarded. This disrupts the statistical consistency of the latent noise, leading to a deviation from the expected real Gaussian distribution and potential degradation in generative quality.
In contrast, embedding with Hermitian symmetry better retains the statistical structure of the latent noise, ensuring that the transformed spatial-domain signal remains closer to a real Gaussian distribution. By preserving these statistical properties, the initialization process in diffusion models becomes more stable, ultimately enhancing generative performance, as observed in \cref{sec:exp-baselines}.

%% file: sec/4_methods.tex
\section{Methods}
\subsection{Hermitian Symmetric Fourier Watermark}
The proposed Hermitian SFW refers to a watermark designed to satisfy the Hermitian symmetry introduced in the previous section. As shown in the left side of \cref{fig:concept}, SFW allows free usage of the half-region in the frequency domain, while the remaining half is restricted by the symmetry condition. Certain constraints must be satisfied when embedding a pattern in the free half-region under a semantic watermarking scheme: the imaginary part at the DC center must be zero, and if the signal dimensions are even, the imaginary components of the points corresponding to the Nyquist frequency must also be zero. In cases where both frequency axes have even dimensions as in this study, the points where the imaginary part must be zero are $(0,0)$, $(\frac{M}{2},0)$, $(0,\frac{N}{2})$, and $(\frac{M}{2},\frac{N}{2})$. Instead of directly modifying the watermark pattern to comply with locations where the imaginary part must be zero, a more effective strategy is to embed it while avoiding the DC axis as much as possible, as will be introduced in \cref{sec:method-HSQR}

\subsection{Center-Aware Embedding Strategy}
Existing semantic watermarking baselines~\cite{wen2024tree, ci2024ringid, zhang2025attack} embed patterns by applying the Fourier transform to the full spatial matrix of the latent vector. However, such methods are vulnerable to spatial attacks, particularly cropping, which can lead to the loss of watermark information and reduced detection performance.
To address this issue, we propose a center-aware embedding strategy, which applies the Fourier transform only to the central area of the spatial domain before embedding. Specifically, for a latent vector with a signal dimension of 64, we utilize the central $44 \times 44$ space. As demonstrated in \cref{sec:ablation-crop}, this design significantly improves robustness against cropping attacks on various scales.

\subsection{Integrating SFW and Center-Aware Design}
\subsubsection{Application to the Baseline}
As shown in \cref{fig:wm-fft-a}, the \textit{Tree-Ring}~\cite{wen2024tree} baseline suffers from frequency loss. To mitigate this, we impose the Hermitian symmetry condition directly on the watermark patterns.
By integrating SFW with the center-aware embedding strategy, we refine the baseline into Hermitian Symmetric Tree-Ring (HSTR). HSTR not only enhances detection performance by fully utilizing the imaginary components of the latent Fourier domain but also improves image generation quality by restoring frequency integrity.

\begin{table*}[th]
\centering
\scriptsize 
\caption{Verification performance of different watermarking methods under various attacks. Bit Accuracy is used for bitstream-based methods (DwtDct, DwtDctSvd, RivaGAN, S.Sign), while TPR@1\%FPR is used for semantic methods. Best performances are highlighted. Our methods show superior detection accuracy and robustness against signal processing distortions, regeneration, and cropping attacks.}
\begin{tabular}{@{}c|l|cccccccccccc|c}
\arrayrulecolor{black}
\toprule
\multirow{2}{*}{\makecell{\\Datasets}} & \multirow{2}{*}{\makecell{\\Methods}} & No Attack &
\multicolumn{6}{c}{Signal Processing Attack} & 
\multicolumn{3}{c}{Regeneration Attack} & 
\multicolumn{2}{c|}{Cropping Attack} &
\multirow{2}{*}{\makecell{\\Avg}} \\
\cmidrule(rl){3-3}
\cmidrule(rl){4-9}
\cmidrule(rl){10-12}
\cmidrule(rl){13-14}
& & Clean & Bright. & Cont. & JPEG & Blur & Noise & BM3D & VAE-B & VAE-C & Diff. & C.C. & R.C. & \\
\midrule
\multirow{9}{*}{\makecell{\\MS-COCO}} 
& DwtDct & 0.863 & 0.572 & 0.522 & 0.516 & 0.677 & 0.859 & 0.532 & 0.523 & 0.521 & 0.519 & 0.729 & 0.810 & 0.637 \\
& DwtDctSvd & \cellcolor{gray!20}{1.000} & 0.555 & 0.473 & 0.602 & \cellcolor{gray!20}{1.000} & \cellcolor{gray!20}{1.000} & 0.784 & 0.648 & 0.596 & 0.644 & 0.744 & 0.861 & 0.742 \\
& RivaGAN & 0.999 & 0.862 & 0.986 & 0.821 & 0.998 & 0.969 & 0.934 & 0.570 & 0.552 & 0.608 & 0.991 & 0.995 & 0.857 \\
& S.Sign. & 0.995 & 0.894 & 0.978 & 0.806 & 0.911 & 0.721 & 0.838 & 0.717 & 0.715 & 0.478 & 0.987 & 0.991 & 0.836 \\
\cmidrule{2-15}
& Tree-Ring & 0.957 & 0.463 & 0.900 & 0.548 & 0.934 & 0.412 & 0.815 & 0.509 & 0.536 & 0.543 & 0.509 & 0.734 & 0.655 \\
& Zodiac & 0.998 & 0.843 & 0.998 & 0.973 & 0.998 & 0.880 & 0.997 & 0.944 & 0.958 & 0.972 & 0.989 & 0.995 & 0.962 \\
& HSTR (ours) & \cellcolor{gray!20}{1.000} & 0.899 & \cellcolor{gray!20}{1.000} & 0.994 & \cellcolor{gray!20}{1.000} & 0.806 & 0.999 & 0.973 & 0.982 & 0.997 & \cellcolor{gray!20}{1.000} & \cellcolor{gray!20}{1.000} & 0.971 \\
\cmidrule{2-15}
& RingID & \cellcolor{gray!20}{1.000} & 0.988 & \cellcolor{gray!20}{1.000} & \cellcolor{gray!20}{1.000} & \cellcolor{gray!20}{1.000} & 0.987 & \cellcolor{gray!20}{1.000} & \cellcolor{gray!20}{0.992} & \cellcolor{gray!20}{1.000} & \cellcolor{gray!20}{1.000} & \cellcolor{gray!20}{1.000} & \cellcolor{gray!20}{1.000} & \cellcolor{gray!20}{0.997} \\
& HSQR (ours) & \cellcolor{gray!20}{1.000} & \cellcolor{gray!20}{0.991} & \cellcolor{gray!20}{1.000} & \cellcolor{gray!20}{1.000} & \cellcolor{gray!20}{1.000} & 0.983 & \cellcolor{gray!20}{1.000} & \cellcolor{gray!20}{0.992} & \cellcolor{gray!20}{1.000} & \cellcolor{gray!20}{1.000} & \cellcolor{gray!20}{1.000} & \cellcolor{gray!20}{1.000} & \cellcolor{gray!20}{0.997} \\

\midrule
\multirow{9}{*}{\makecell{\\SD-Prompts}} 
& DwtDct & 0.819 & 0.557 & 0.516 & 0.506 & 0.685 & 0.822 & 0.530 & 0.513 & 0.512 & 0.509 & 0.723 & 0.794 & 0.624 \\
& DwtDctSvd & \cellcolor{gray!20}{1.000} & 0.537 & 0.459 & 0.610 & 0.999 & \cellcolor{gray!20}{0.998} & 0.859 & 0.659 & 0.620 & 0.623 & 0.743 & 0.860 & 0.747 \\
& RivaGAN & 0.991 & 0.823 & 0.963 & 0.810 & 0.988 & 0.961 & 0.915 & 0.572 & 0.535 & 0.567 & 0.980 & 0.983 & 0.841 \\
& S.Sign. & 0.994 & 0.899 & 0.967 & 0.769 & 0.888 & 0.742 & 0.809 & 0.677 & 0.671 & 0.493 & 0.983 & 0.990 & 0.824 \\
\cmidrule{2-15}
& Tree-Ring & 0.944 & 0.471 & 0.894 & 0.466 & 0.912 & 0.423 & 0.802 & 0.509 & 0.514 & 0.543 & 0.469 & 0.749 & 0.641 \\
& Zodiac & 0.998 & 0.748 & 0.999 & 0.979 & 0.999 & 0.903 & \cellcolor{gray!20}{1.000} & 0.940 & 0.975 & 0.958 & 0.994 & 0.996 & 0.957 \\
& HSTR (ours) & \cellcolor{gray!20}{1.000} & 0.742 & \cellcolor{gray!20}{1.000} & 0.990 & \cellcolor{gray!20}{1.000} & 0.850 & \cellcolor{gray!20}{1.000} & 0.983 & 0.987 & 0.999 & \cellcolor{gray!20}{1.000} & \cellcolor{gray!20}{1.000} & 0.963 \\
\cmidrule{2-15}
& RingID & \cellcolor{gray!20}{1.000} & \cellcolor{gray!20}{0.972} & \cellcolor{gray!20}{1.000} & \cellcolor{gray!20}{1.000} & \cellcolor{gray!20}{1.000} & 0.988 & \cellcolor{gray!20}{1.000} & \cellcolor{gray!20}{0.996} & \cellcolor{gray!20}{1.000} & \cellcolor{gray!20}{1.000} & \cellcolor{gray!20}{1.000} & \cellcolor{gray!20}{1.000} & \cellcolor{gray!20}{0.996} \\
& HSQR (ours) & \cellcolor{gray!20}{1.000} & 0.955 & \cellcolor{gray!20}{1.000} & 0.999 & \cellcolor{gray!20}{1.000} & 0.992 & \cellcolor{gray!20}{1.000} & \cellcolor{gray!20}{0.996} & 0.999 & \cellcolor{gray!20}{1.000} & \cellcolor{gray!20}{1.000} & \cellcolor{gray!20}{1.000} & 0.995 \\

\midrule
\multirow{9}{*}{\makecell{\\DiffusionDB}} 
& DwtDct & 0.842 & 0.563 & 0.515 & 0.509 & 0.672 & 0.829 & 0.526 & 0.513 & 0.514 & 0.512 & 0.723 & 0.801 & 0.627 \\
& DwtDctSvd & 0.998 & 0.558 & 0.463 & 0.593 & 0.997 & \cellcolor{gray!20}{0.995} & 0.830 & 0.658 & 0.608 & 0.621 & 0.742 & 0.860 & 0.744 \\
& RivaGAN & 0.987 & 0.839 & 0.960 & 0.790 & 0.985 & 0.937 & 0.893 & 0.553 & 0.518 & 0.556 & 0.974 & 0.979 & 0.831 \\
& S.Sign. & 0.990 & 0.890 & 0.967 & 0.787 & 0.889 & 0.726 & 0.819 & 0.690 & 0.687 & 0.496 & 0.981 & 0.986 & 0.826 \\
\cmidrule{2-15}
& Tree-Ring & 0.940 & 0.487 & 0.889 & 0.434 & 0.904 & 0.392 & 0.799 & 0.454 & 0.503 & 0.454 & 0.499 & 0.715 & 0.622 \\
& Zodiac & 0.992 & 0.752 & 0.988 & 0.933 & 0.988 & 0.834 & 0.984 & 0.911 & 0.926 & 0.903 & 0.971 & 0.985 & 0.931 \\
& HSTR (ours) & 0.999 & 0.792 & 0.996 & 0.981 & 0.996 & 0.792 & 0.991 & 0.968 & 0.969 & 0.989 & \cellcolor{gray!20}{1.000} & \cellcolor{gray!20}{1.000} & 0.956 \\
\cmidrule{2-15}
& RingID & \cellcolor{gray!20}{1.000} & \cellcolor{gray!20}{0.989} & \cellcolor{gray!20}{1.000} & \cellcolor{gray!20}{1.000} & \cellcolor{gray!20}{1.000} & 0.963 & \cellcolor{gray!20}{1.000} & 0.995 & \cellcolor{gray!20}{0.999} & \cellcolor{gray!20}{1.000} & \cellcolor{gray!20}{1.000} & \cellcolor{gray!20}{1.000} & \cellcolor{gray!20}{0.995} \\
& HSQR (ours) & \cellcolor{gray!20}{1.000} & 0.977 & \cellcolor{gray!20}{1.000} & 0.999 & \cellcolor{gray!20}{1.000} & 0.974 & 0.999 & \cellcolor{gray!20}{0.997} & \cellcolor{gray!20}{0.999} & \cellcolor{gray!20}{1.000} & \cellcolor{gray!20}{1.000} & \cellcolor{gray!20}{1.000} & \cellcolor{gray!20}{0.995} \\
\bottomrule
\end{tabular}
\label{tab:verify-t@1f-ba}
\end{table*}

\subsubsection{Hermitian Symmetric QR code}
\label{sec:method-HSQR}
In this section, we introduce a novel approach that extends SFW beyond baselines to QR codes, which are widely recognized for their high versatility and robust error correction capabilities~\cite{denso_qr_code}.

\noindent
\textbf{Embedding.}
As illustrated in \cref{fig:wm-fft-b}, the Hermitian Symmetric QR Code (HSQR) watermark is constructed by splitting the QR code in half and embedding each part separately into the real and imaginary components of the free half-region in the Fourier domain. The binary pattern of the QR code is embedded using the following formulation:

\begin{equation}
\operatorname{HSQR}(\tilde{x}, c) = 
\begin{cases} 
+|F(\tilde{x}, c)|, & \text{if } \operatorname{QR}(x) = 1 \\
-|F(\tilde{x}, c)|, & \text{if } \operatorname{QR}(x) = 0 ,
\end{cases}
\end{equation}
where: 
\begin{itemize}
    \item $x$ denotes the coordinates in the QR code domain.
    \item $\tilde{x}$ 
    represents the corresponding coordinates in the embedding region of the Fourier domain. Since the QR code is split into two halves, each half is mapped to either the real or imaginary component.
    \item $c \in \{\text{Re}, \text{Im}\}$ indicates whether the embedding occurs in the real or imaginary part.
    \item $F(\tilde{x}, c)$ is the real-valued amplitude of the complex Gaussian noise in the Fourier domain, where a complex number is expressed as \( F_{\text{Re}} + iF_{\text{Im}} \), making both real and imaginary amplitudes individually real-valued.
\end{itemize}
To maintain symmetry while avoiding numerical instability, the embedding region is positioned one pixel to the right of the vertical DC axis.
Additionally, to increase the amount of embedded information and statistically reduce errors, each QR code cell is represented by multiple pixels arranged in a square pattern, e.g., $2 \times 2$ pixels.

\noindent
\textbf{Detection.}
The detection process for semantic watermarking is performed by computing the L1 distance. In HSQR, the ground truth QR binary pattern is converted into a signed Boolean representation as follows:
\begin{equation}
\operatorname{QR^*}(\tilde{x}, c) = 
\begin{cases} 
+\Lambda, & \text{if } \operatorname{QR}(x) = 1 \\
-\Lambda, & \text{if } \operatorname{QR}(x) = 0 ,
\end{cases}
\end{equation}
where $\Lambda$ is a fixed amplitude for binary encoding.


%% file: sec/5_experiments.tex
\begin{table*}[th]
\centering
\scriptsize 
\caption{Identification accuracy for different watermarking methods, evaluated across multiple attack conditions. Perfect Match Rate is used for bitstream-based methods. The best performance for each item is highlighted with shading. HSQR achieves state-of-the-art performance, while HSTR consistently outperforms other Gaussian radius-based semantic baselines, particularly under cropping attacks.}
\begin{tabular}{@{}c|l|cccccccccccc|c}
\arrayrulecolor{black}
\toprule
\multirow{2}{*}{\makecell{\\Datasets}} & \multirow{2}{*}{\makecell{\\Methods}} & No Attack &
\multicolumn{6}{c}{Signal Processing Attack} & 
\multicolumn{3}{c}{Regeneration Attack} & 
\multicolumn{2}{c|}{Cropping Attack} &
\multirow{2}{*}{\makecell{\\Avg}} \\
\cmidrule(rl){3-3}
\cmidrule(rl){4-9}
\cmidrule(rl){10-12}
\cmidrule(rl){13-14}
& & Clean & Bright. & Cont. & JPEG & Blur & Noise & BM3D & VAE-B & VAE-C & Diff. & C.C. & R.C. & \\
\midrule
\multirow{9}{*}{\makecell{\\MS-COCO}} 
& DwtDct & 0.466 & 0.044 & 0.000 & 0.000 & 0.038 & 0.442 & 0.000 & 0.000 & 0.000 & 0.000 & 0.000 & 0.000 & 0.083 \\
& DwtDctSvd & \cellcolor{gray!20}{1.000} & 0.044 & 0.019 & 0.000 & 0.999 & \cellcolor{gray!20}{0.998} & 0.037 & 0.000 & 0.000 & 0.003 & 0.000 & 0.000 & 0.258 \\
& RivaGAN & 0.974 & 0.260 & 0.772 & 0.023 & 0.961 & 0.686 & 0.348 & 0.000 & 0.000 & 0.000 & 0.852 & 0.909 & 0.482 \\
& S.Sign. & 0.873 & 0.177 & 0.563 & 0.000 & 0.036 & 0.010 & 0.007 & 0.000 & 0.000 & 0.000 & 0.709 & 0.802 & 0.265 \\
\cmidrule{2-15}
& Tree-Ring & 0.303 & 0.087 & 0.207 & 0.072 & 0.256 & 0.030 & 0.162 & 0.083 & 0.072 & 0.054 & 0.009 & 0.033 & 0.114 \\
& Zodiac & 0.000 & 0.000 & 0.000 & 0.000 & 0.000 & 0.000 & 0.000 & 0.000 & 0.000 & 0.000 & 0.000 & 0.000 & 0.000 \\
& HSTR (ours) & \cellcolor{gray!20}{1.000} & 0.714 & 0.999 & 0.886 & 0.998 & 0.460 & 0.972 & 0.833 & 0.831 & 0.971 & \cellcolor{gray!20}{1.000} & \cellcolor{gray!20}{1.000} & 0.889 \\
\cmidrule{2-15}
& RingID & \cellcolor{gray!20}{1.000} & 0.875 & \cellcolor{gray!20}{1.000} & 0.975 & \cellcolor{gray!20}{1.000} & 0.919 & 0.996 & 0.978 & 0.970 & 0.998 & 0.874 & 0.978 & 0.964 \\
& HSQR (ours) & \cellcolor{gray!20}{1.000} & \cellcolor{gray!20}{0.958} & \cellcolor{gray!20}{1.000} & \cellcolor{gray!20}{0.994} & \cellcolor{gray!20}{1.000} & 0.901 & \cellcolor{gray!20}{0.999} & \cellcolor{gray!20}{0.980} & \cellcolor{gray!20}{0.987} & \cellcolor{gray!20}{0.999} & \cellcolor{gray!20}{1.000} & \cellcolor{gray!20}{1.000} & \cellcolor{gray!20}{0.985} \\

\midrule
\multirow{9}{*}{\makecell{\\SD-Prompts}} 
& DwtDct & 0.285 & 0.024 & 0.000 & 0.000 & 0.017 & 0.276 & 0.000 & 0.000 & 0.000 & 0.000 & 0.000 & 0.000 & 0.050 \\
& DwtDctSvd & 0.993 & 0.028 & 0.011 & 0.000 & 0.982 & \cellcolor{gray!20}{0.979} & 0.085 & 0.000 & 0.000 & 0.007 & 0.000 & 0.000 & 0.257 \\
& RivaGAN & 0.878 & 0.213 & 0.613 & 0.009 & 0.857 & 0.657 & 0.304 & 0.000 & 0.000 & 0.000 & 0.736 & 0.768 & 0.420 \\
& S.Sign. & 0.813 & 0.263 & 0.420 & 0.000 & 0.021 & 0.015 & 0.006 & 0.000 & 0.000 & 0.000 & 0.576 & 0.716 & 0.236 \\
\cmidrule{2-15}
& Tree-Ring & 0.288 & 0.094 & 0.189 & 0.051 & 0.235 & 0.034 & 0.159 & 0.079 & 0.076 & 0.056 & 0.012 & 0.041 & 0.110 \\
& Zodiac & 0.000 & 0.000 & 0.000 & 0.000 & 0.000 & 0.000 & 0.000 & 0.000 & 0.000 & 0.000 & 0.000 & 0.000 & 0.000 \\
& HSTR (ours) & \cellcolor{gray!20}{1.000} & 0.655 & 0.999 & 0.863 & 0.999 & 0.555 & 0.980 & 0.846 & 0.847 & 0.973 & \cellcolor{gray!20}{1.000} & \cellcolor{gray!20}{1.000} & 0.893 \\
\cmidrule{2-15}
& RingID & \cellcolor{gray!20}{1.000} & 0.885 & \cellcolor{gray!20}{1.000} & 0.976 & 0.998 & 0.886 & 0.993 & 0.980 & 0.973 & 0.995 & 0.876 & 0.981 & 0.962 \\
& HSQR (ours) & \cellcolor{gray!20}{1.000} & \cellcolor{gray!20}{0.930} & \cellcolor{gray!20}{1.000} & \cellcolor{gray!20}{0.994} & \cellcolor{gray!20}{1.000} & 0.942 & \cellcolor{gray!20}{0.999} & \cellcolor{gray!20}{0.991} & \cellcolor{gray!20}{0.997} & \cellcolor{gray!20}{1.000} & \cellcolor{gray!20}{1.000} & \cellcolor{gray!20}{1.000} & \cellcolor{gray!20}{0.988} \\

\midrule
\multirow{9}{*}{\makecell{\\DiffusionDB}} 
& DwtDct & 0.357 & 0.037 & 0.000 & 0.000 & 0.034 & 0.320 & 0.000 & 0.000 & 0.000 & 0.000 & 0.000 & 0.000 & 0.062 \\
& DwtDctSvd & 0.990 & 0.036 & 0.019 & 0.000 & 0.975 & \cellcolor{gray!20}{0.959} & 0.081 & 0.000 & 0.000 & 0.001 & 0.000 & 0.000 & 0.255 \\
& RivaGAN & 0.858 & 0.213 & 0.625 & 0.020 & 0.848 & 0.615 & 0.221 & 0.000 & 0.000 & 0.000 & 0.756 & 0.780 & 0.411 \\
& S.Sign. & 0.798 & 0.207 & 0.472 & 0.000 & 0.027 & 0.005 & 0.005 & 0.000 & 0.000 & 0.000 & 0.643 & 0.738 & 0.241 \\
\cmidrule{2-15}
& Tree-Ring & 0.280 & 0.095 & 0.190 & 0.059 & 0.233 & 0.037 & 0.145 & 0.081 & 0.072 & 0.050 & 0.013 & 0.039 & 0.108 \\
& Zodiac & 0.000 & 0.000 & 0.000 & 0.000 & 0.000 & 0.000 & 0.000 & 0.000 & 0.000 & 0.000 & 0.000 & 0.000 & 0.000 \\
& HSTR (ours) & 0.996 & 0.721 & 0.992 & 0.854 & 0.989 & 0.563 & 0.958 & 0.830 & 0.821 & 0.952 & 0.996 & 0.996 & 0.889 \\
\cmidrule{2-15}
& RingID & \cellcolor{gray!20}{1.000} & 0.895 & \cellcolor{gray!20}{1.000} & 0.947 & 0.996 & 0.871 & 0.992 & 0.968 & 0.958 & 0.990 & 0.875 & 0.984 & 0.956 \\
& HSQR (ours) & \cellcolor{gray!20}{1.000} & \cellcolor{gray!20}{0.954} & \cellcolor{gray!20}{1.000} & \cellcolor{gray!20}{0.988} & \cellcolor{gray!20}{1.000} & 0.906 & \cellcolor{gray!20}{0.998} & \cellcolor{gray!20}{0.982} & \cellcolor{gray!20}{0.991} & \cellcolor{gray!20}{0.994} & \cellcolor{gray!20}{1.000} & \cellcolor{gray!20}{1.000} & \cellcolor{gray!20}{0.984} \\
\bottomrule
\end{tabular}
\label{tab:identify}
\end{table*}

\section{Experiments}
\subsection{Experimental Settings}
\noindent
\textbf{Semantic Watermarks.} The \textit{Tree-Ring}~\cite{wen2024tree} watermark is embedded into channel 3 with a radius of 14 and HSTR adopts the same watermark parameters as its baseline. For \textit{RingID}~\cite{ci2024ringid}, we follow the original settings, embedding a tree-ring pattern with a radius range of 3-14 in channel 3, along with a Gaussian noise key in channel 0. Both SFW methods follow \textit{RingID} in utilizing a Gaussian noise key in channel 0. For HSQR, we use a version 1 QR code ($21 \times 21$ cells) with a cell size of 2 pixels and error correction level H, capable of encoding up to 72 bits. The QR code is embedded in channel 3, and the encoding amplitude $\Lambda$ is set to 45, corresponding to the standard deviation of the real and imaginary components in the Fourier domain of $64 \times 64$ normal Gaussian latent vector ($\approx\sqrt{64^2/2}$). Note that these experiments are conducted on $512 \times 512$ images. At higher resolutions, a larger embedding space allows for increased capacity through higher QR code versions that encode more bits, as well as the use of stronger watermarking parameters, enabling greater robustness and adaptability. Additionally, the post-hoc semantic watermarking method, Zodiac~\cite{zhang2025attack}, is included in our comparative experiments.

\noindent
\textbf{Model Setup and Datasets.} The diffusion model used for the experiments is Stable Diffusion v2-1-base~\cite{rombach2022high}, configured to generate images at a resolution of $512 \times 512$ with a CFG scale of $7.5$. Both generation and inversion steps of the DDIM scheduler~\cite{song2020denoising} are set to 50.
The datasets used for image generation include 5,000 captions from the MS-COCO-2017 training set~\cite{lin2014microsoft}, 1,001 sampled prompts from the 2M subset of DiffusionDB~\cite{wang2022diffusiondb}, and 8,192 prompts from the Stable-Diffusion-Prompts test set~\cite{gustavo}. For watermark detection, the number of watermarked images used for each dataset is set to 1,000. Specifically, verification is conducted on 1,000 pairs of watermarked and unwatermarked images, while identification is performed on 1,000 watermarked images.

\noindent
\textbf{Attack Methods.} 
To evaluate the robustness of our method against various distortions, we apply 11 different attacks, categorized into three major types.
\begin{itemize}
    \item Signal Processing Attacks: These include brightness adjustment (ranging from 0 to 7), contrast adjustment (factor of 0.5), JPEG compression (quality factor of 25), Gaussian blur (radius of 5), Gaussian noise ($\sigma=0.05$), and BM3D denoising ($\sigma=0.1$).
    \item Regeneration Attacks: We apply two VAE-based image compression models, Bmshj18~\cite{balle2018variational} (VAE-B) and Cheng20~\cite{cheng2020learned} (VAE-C), both at quality level 3, as well as a diffusion-based regeneration attack by Zhao \etal~\cite{zhao2025invisible} (referred to as Diff.) using 60 denoising steps.
    \item Cropping Attacks: We apply center crop (C.C.) with a crop scale of 0.5 and random crop (R.C.) with a crop scale of 0.7, where the crop scale represents the ratio of the cropped image area to the original image.
\end{itemize}

\noindent
\textbf{Evaluation Metrics.} 
The $L_{1}$ distance is computed only within the key region of the complex Fourier domain of latent noise, and unless otherwise specified, 2,048 keys are used for identification accuracy.
For verification, semantic watermarking methods use True Positive Rate at 1\% False Positive Rate (TPR@1\%FPR) as the evaluation metric. To ensure a fair comparison with bitstream-based approaches such as DwtDct, DwtDctSvd~\cite{cox2007digital}, RivaGAN~\cite{zhang2019robust}, and Stable Signature (S.Sign.)~\cite{fernandez2023stable}, we measure Bit Accuracy for these methods. For identification, where the goal is to match the exact message, Perfect Match Rate is used as the metric for bitstream-based approaches. To evaluate the generation quality, FID~\cite{heusel2017gans} is measured between watermarked images and 5,000 MS-COCO ground truth images. We also calculate the CLIP score~\cite{radford2021learning} using the OpenCLIP-ViT/G model~\cite{cherti2023reproducible} to determine how closely the generated image aligns with the input prompt.

\begin{table}[t]
\centering
\scriptsize 
\caption{Generative quality evaluation of watermarking methods based on FID (MS-COCO ground truth) and CLIP score. The best performance for each item is highlighted with shading, while bold text specifically marks the low CLIP score in RingID. Our proposed methods preserve frequency integrity, achieving the best balance between watermark robustness and generative performance, whereas RingID introduces visible artifacts, compromising perceptual quality. \textit{Vrf.} and \textit{Idf.} denote the average detection performance in verification and identification tasks, respectively.}
\begin{tabular}{l|l|cc|cc}
\toprule
\multicolumn{2}{c|}{Semantic Methods} & FID $\downarrow$ & CLIP $\uparrow$ & \textit{Vrf.} & \textit{Idf.}\\
\midrule
\multirow{4}{*}{\makecell{\\Merged in\\Generation}} 
& Tree-Ring & 26.418 & 0.326 & 0.655 & 0.114 \\
& RingID & 27.052 & \textbf{0.324} & \cellcolor{gray!20}{0.997} & 0.964 \\
\cmidrule{2-6}
& HSTR (ours) & 25.062 & 0.329 & 0.971 & 0.889\\
& HSQR (ours) & \cellcolor{gray!20}{24.895} & \cellcolor{gray!20}{0.330} & \cellcolor{gray!20}{0.997} & \cellcolor{gray!20}{0.985} \\
\bottomrule
\end{tabular}
\label{tab:gen_quality}
\end{table}

\subsection{Comparison with Baselines}
\label{sec:exp-baselines}
In this section, we compare the detection performance in both verification and identification tasks, as well as the generation quality of different watermarking methods.

\noindent
\textbf{Enhanced Detection Robustness.}
According to \cref{tab:verify-t@1f-ba}, our proposed methods consistently achieve top-tier detection performance across various attacks. A key observation is that non-semantic watermarking methods~\cite{cox2007digital, zhang2019robust, fernandez2023stable} exhibit significant degradation under regeneration attacks, highlighting their vulnerability.
Among Gaussian radius-based pattern methods (Tree-Ring, Zodiac, and HSTR), HSTR outperforms corresponding baselines in most cases, demonstrating its effectiveness in enhancing detection robustness while maintaining the fundamental tree-ring structure. In contrast, RingID leverages high-energy signed constant ring patterns, securing strong detection performance. However, as acknowledged by the authors, this method introduces noticeable \textit{ring-like} artifacts in generated images, disrupting the balance between watermark robustness and generative quality. A detailed quantitative evaluation of this trade-off is presented in the following section.
Meanwhile, Zodiac suffers from time-consuming processing, taking several minutes to apply the watermark---requiring 7.36 minutes per image on MS-COCO dataset---which further limits its practicality in real-world applications. In contrast, our methods following the \textit{merged-in-generation} scheme introduce no additional processing time, ensuring seamless watermark embedding.

In \cref{tab:identify}, we compare identification results, which reveal an even greater disparity in detection performance across different watermarking methods. Gaussian radius-based pattern methods, except for HSTR, perform poorly in identification tasks, with Zodiac---designed solely for verification---failing entirely across all scenarios (achieving zero identification accuracy). Further analysis on scalability under different message capacities is provided in \cref{sec:ablation-capacity}.
Notably, our methods exhibit strong resilience against cropping attacks in both tasks, reinforcing the robustness of center-aware embedding. HSQR achieves state-of-the-art identification accuracy across all datasets, further establishing its dominance in watermark retrieval.

\noindent
\textbf{Balance with Generative Quality.}
\cref{tab:gen_quality} presents the generative quality of different semantic watermarking methods following the \textit{merged-in-generation} scheme, evaluated using FID and CLIP score. RingID, which deviates from a Gaussian distribution by embedding high-energy perturbations, exhibits the worst generative performance, particularly reflected in its low CLIP score. This decline is attributed to the noticeable ring-like artifacts, as discussed earlier.
In contrast, our proposed methods, which preserve frequency integrity via SFW, achieve the top two FID scores, demonstrating a better balance between watermark robustness and generative quality.

\noindent
\textbf{Expanding Frequency Utilization in Latent Watermarking.}
Traditional image-domain watermarking methods prioritize low-mid frequency embedding to resist compression and filtering attacks. This approach has also been adopted in latent diffusion-based semantic watermarking methods~\cite{wen2024tree, ci2024ringid, zhang2025attack} without explicitly considering the differences between pixel-space and latent-space transformations. However, our findings suggest that such frequency constraints may not be necessary in the latent space, where watermark retrieval involves latent encoding and DDIM inversion, altering the impact of frequency perturbations.
This distinction is evident in HSQR, which achieves state-of-the-art detection performance while utilizing nearly the entire frequency spectrum. Instead of relying on specific frequency bands, HSQR leverages structured statistical encoding, distributing watermark information across multiple pixels ($2 \times 2$ per QR cell). This redundancy enhances robustness against distortions, enabling effective retrieval even with broader frequency usage.
These results indicate that latent-space watermarking is not bounded by traditional low-mid frequency constraints. Future semantic watermarking strategies can benefit from statistically robust encoding methods, allowing for effective watermark retrieval across a wider frequency range without compromising detection accuracy.

\begin{table}[t]
\centering
\scriptsize
\caption{Ablation study on detection performance based on frequency integrity (\(\checkmark\) or \(\times\)) and the number of detection region usage (1: real only, 2: real \& imaginary). \textit{Vrf.} and \textit{Idf.} represent average detection performance in verification (TPR@1\% FPR) and identification (accuracy), respectively. $\Delta L_1^*$ denotes the normalized $\Delta L_1$ metric, indicating detection effectiveness.}
\begin{tabular}{c|lcc|cc|c}
\toprule
Case & Methods & Freq. Int. & \# Det. & \textit{Vrf.} & \textit{Idf.} & $\Delta L_1^* \uparrow$ \\
\midrule
A & Tree-Ring & \(\times\) & 2 & 0.653 & 0.114 & 0.232 \\
B & Tree-Ring & \(\times\) & 1 & 0.805 & 0.416 & 0.368 \\
C & HSTR (ours) & \(\checkmark\) & 1 & 0.936 & 0.775 & 0.471 \\
D & HSTR (ours) & \(\checkmark\) & 2 & 0.971 & 0.889 & 0.476 \\
\bottomrule
\end{tabular}
\label{tab:ablation-scope}
\end{table}

\begin{figure}[t]
\centering
\includegraphics[width=\linewidth]{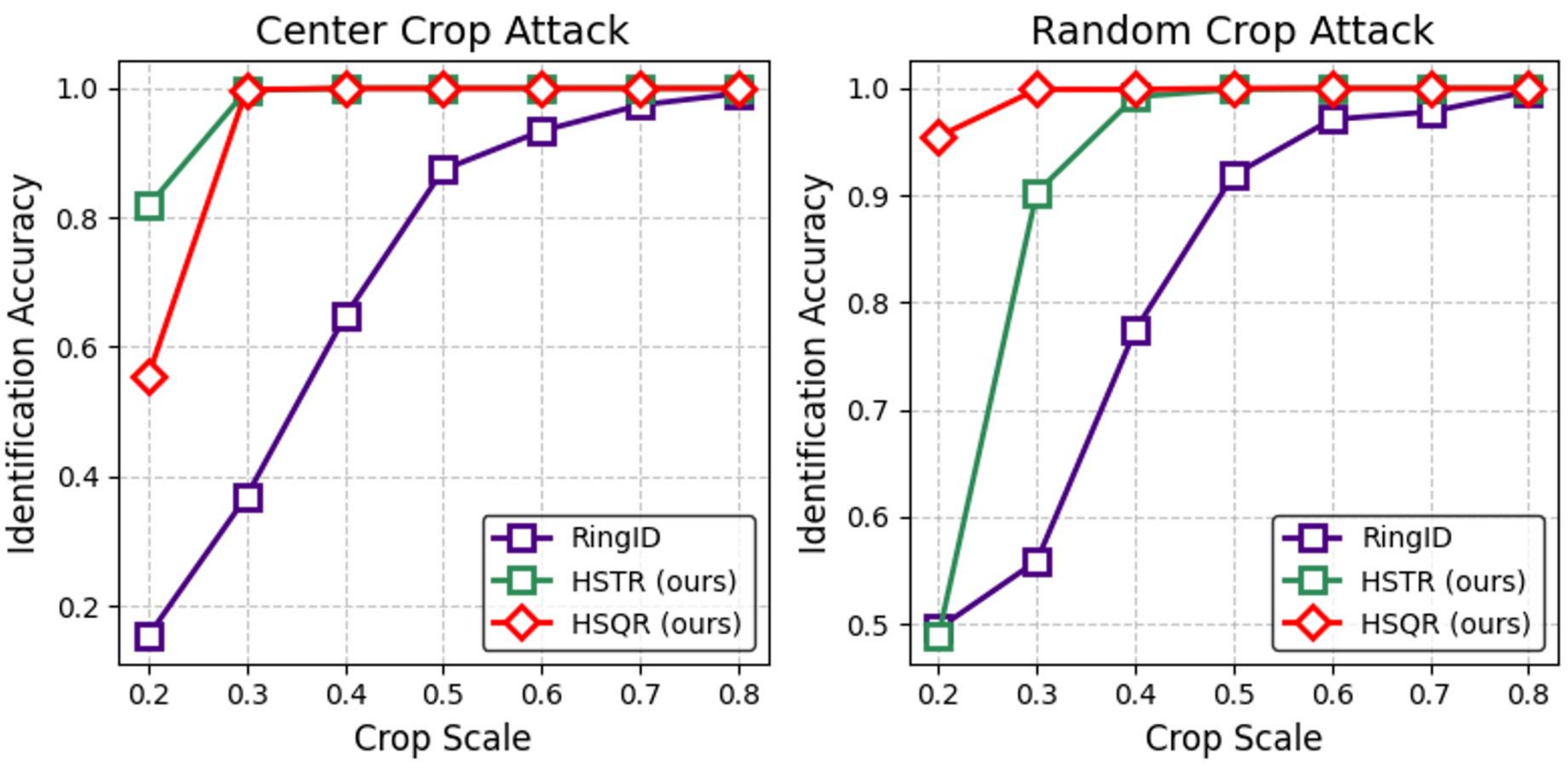}
\caption{Identification accuracy under center crop and random crop attacks at different crop scales. HSTR and HSQR maintain higher accuracy compared to RingID, demonstrating improved robustness against cropping.}
\label{fig:ablation-crop}
\end{figure}

\begin{figure}[t]
\centering
\includegraphics[width=\linewidth]{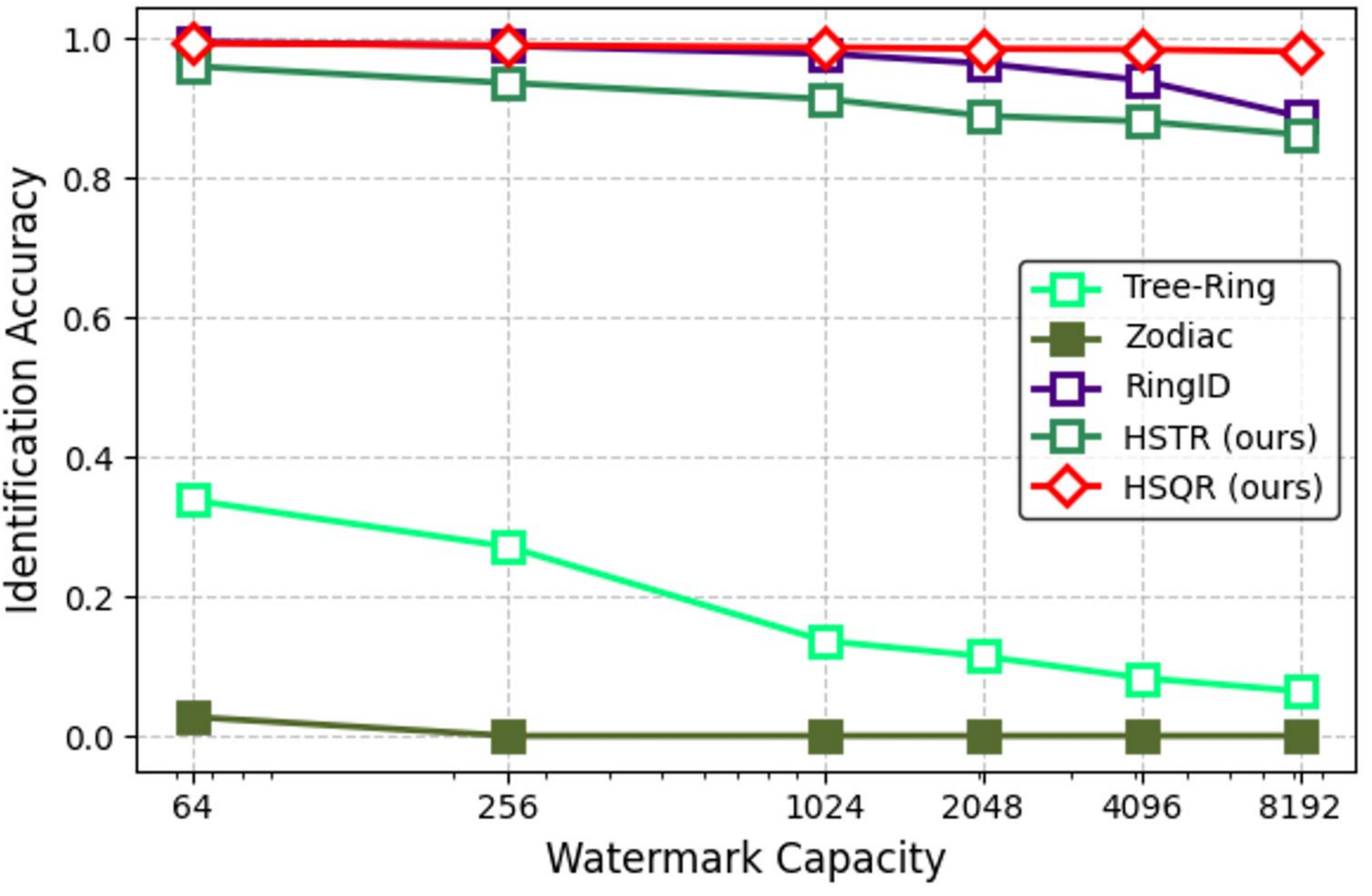}
\caption{Identification accuracy across watermark message capacities. HSQR remains nearly perfect, while RingID and HSTR degrade at higher capacities. Tree-Ring and Zodiac fail to scale effectively.}
\label{fig:ablation-capacity}
\end{figure}

\subsection{Ablation Study}
\label{sec:ablations}
\subsubsection{Impact of SFW on Detection Performance}
\label{sec:ablation-scope}
We evaluate the impact of Hermitian SFW by comparing detection performance across four cases (\cref{tab:ablation-scope}). Without SFW, using both real and imaginary components for detection (Case A) results in the lowest performance due to frequency degradation, while restricting detection to the real component (Case B) improves accuracy. In contrast, applying SFW (Cases C and D) preserves frequency integrity, leading to a significant boost in detection performance. Notably, HSTR with SFW and full complex detection (Case D) achieves the highest accuracy, demonstrating that leveraging both frequency components maximizes retrieval effectiveness. These results confirm that SFW enables robust semantic watermarking by maintaining frequency integrity and fully utilizing the Fourier domain for detection.

\subsubsection{Robustness to Cropping Attacks}
\label{sec:ablation-crop}
We assess the robustness of center-aware embedding against center crop and random crop attacks, measuring identification accuracy across different crop scales (\cref{fig:ablation-crop}). As crop scale decreases, accuracy declines for all methods. However, RingID exhibits a steeper drop, indicating higher vulnerability to cropping. In contrast, HSTR and HSQR degrade more gradually, demonstrating improved resilience. While extreme cropping (scale 0.2) significantly affects all methods, our center-aware design consistently outperforms RingID, confirming its effectiveness in preserving watermark information under cropping distortions.

\subsubsection{Impact of Capacity on Identification}
\label{sec:ablation-capacity}
We evaluate identification accuracy across different watermark message capacities (64 to 8,192) for five methods (\cref{fig:ablation-capacity}). Zodiac's performance collapses, while Tree-Ring exhibits rapid performance degradation, becoming nearly unusable at higher capacities. RingID, HSTR, and HSQR remain robust, maintaining 80\%+ accuracy, though HSTR declines faster, and RingID starts degrading from 2,048 onward. HSQR demonstrates the highest scalability, retaining near-perfect accuracy even at the largest capacity, confirming its superior robustness in high-capacity scenarios.

%% file: sec/6_conclusion.tex
\section{Conclusion}
We have introduced Hermitian SFW, a novel approach to semantic watermarking in the latent diffusion model framework. Unlike existing methods that fail to preserve frequency integrity, our approach ensures that watermark embeddings maintain a consistent statistical structure in the latent noise distribution. This is achieved through Hermitian symmetry enforcement, which preserves frequency components and enhances detection and generation quality.
Additionally, we have proposed center-aware embedding, which significantly improves robustness against cropping attacks by strategically placing watermarks in a spatially resilient region of the latent representation. Through comprehensive experiments, we demonstrated that our method achieves state-of-the-art detection accuracy in both verification and identification tasks while also maintaining superior image generation quality, as shown by FID and CLIP scores.
Our study highlights the importance of frequency integrity in Fourier-based watermarking and challenges the assumption that semantic watermarking must be confined to low-mid frequency bands. Experimental results confirm that a properly structured frequency-domain watermark can be effectively embedded and retrieved across the entire frequency spectrum without compromising generative quality.
Future work includes exploring adaptive embedding strategies to further enhance robustness against adversarial attacks and extreme distortions, as well as extending our method to more diverse generative architectures beyond LDMs.

%% file: sec/X_suppl.tex
\clearpage
\setcounter{page}{1}
\maketitlesupplementary

This supplementary document provides additional context, experiments, and analyses to complement the main paper. 
\cref{sec_supp:task-overview} clarifies the task setup and addresses points of potential misunderstanding.
\cref{sec_supp:evidence} presents additional experimental evidence that reinforces our claims, including new results added in response to reviewer feedback.  
\cref{sec_supp:supplements} provides supplementary quantitative results and extended evaluations with alternative metrics.
\cref{sec_supp:deployment} concludes with a discussion on real-world deployment considerations, highlighting the compatibility of our methods with efficient AI accelerators such as Neural Processing Units (NPUs).

\section{Clarifications and Task Overview}
\label{sec_supp:task-overview}
\subsection{Scope Clarification on Tampering Robustness}
Our method is designed for robust watermarking. It aims to preserve the embedded information even when the content undergoes typical, non-malicious changes during distribution or transformation. 
It is not intended to detect peripheral tampering, which presents a fundamentally different challenge outside the scope of this work. Such tampering detection requires an alternative threat model and design considerations, often involving explicit modeling of adversarial behavior. We clarify this distinction to prevent misunderstanding regarding the intended threat model and design goals of our approach.

\subsection{Taxonomy of Watermarking Methods}
This section provides a brief overview of the terminology used to categorize watermarking methods discussed in the main paper. These categorizations help clarify the design characteristics of each method and contextualize the experimental results.

We group watermarking methods along the following three axes:
\begin{itemize}
    \item \textbf{Message Type:} Methods are either \textit{bitstream-based}, which embed and recover discrete bit sequences, or \textit{pattern-based}, where detection relies on matching structured watermark patterns.
    \item \textbf{Embedding Strategy:} \textit{Post-hoc-based} methods embed watermarks after image generation. In contrast, \textit{merged-in-generation} methods integrate watermarking into the image synthesis process, typically within diffusion-based models.
    \item \textbf{Method Family:} We use the terms \textit{classical vision} for signal processing techniques, and \textit{deep learning-based} for methods involving trainable models. \textit{Semantic watermarking} refers to recent approaches that embed information into the image's semantic content, often in the latent space of generative models. Semantic methods are particularly designed to be robust against semantic-preserving transformations such as regeneration, compression, or cropping.
\end{itemize}

Among the semantic watermarking methods, we further differentiate the structure of their watermark patterns. Specifically, \textit{Gaussian radius-based patterns}, such as Tree-Ring and HSTR, use radial embeddings with Gaussian-sampled values, whereas \textit{structured binary patterns}, such as RingID and HSQR, resemble geometric encodings of bitstreams. 
Although not all of these terms are explicitly mentioned in the main paper, we include them here to help clarify the conceptual distinctions among recent semantic watermarking methods.

\begin{figure}[t]
\centering
\includegraphics[width=0.9\linewidth]{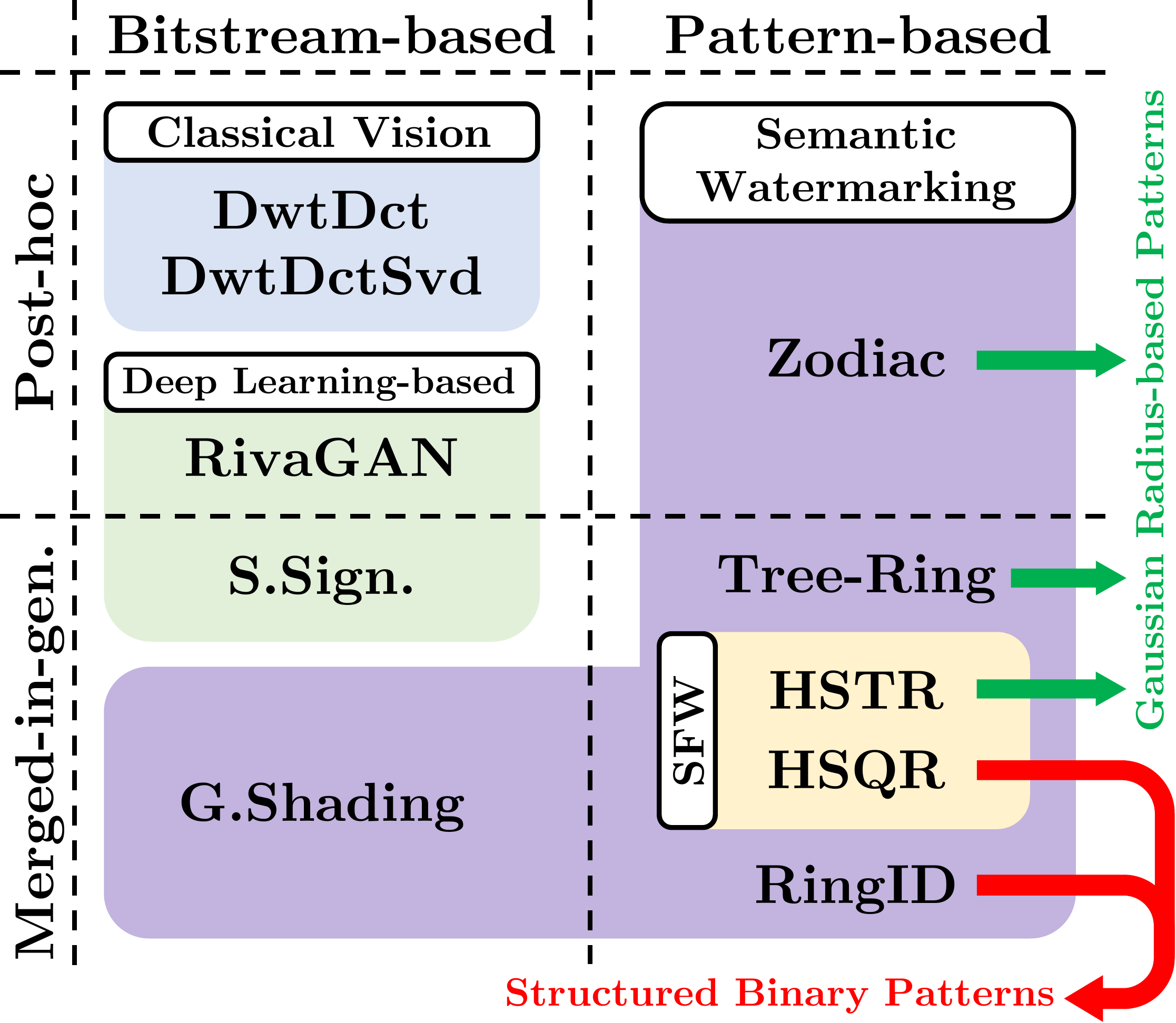}
\caption{Taxonomy of watermarking methods evaluated in this work, categorized along three dimensions.}
\label{fig:supple-terminology}
\end{figure}

\cref{fig:supple-terminology} visually summarizes the classification of all methods evaluated, including the baseline that will be introduced in \cref{sec_supp:G.shading}.

\subsection{Identification Protocol for Tree-Ring}
This section outlines the identification procedure used for the Tree-Ring baseline.
Following the multi-key evaluation protocol introduced in RingID~\cite{ci2024ringid}, we construct a candidate key pool for each target capacity. Specifically, we generate a large set of key embeddings corresponding to different watermark messages. During evaluation, the extracted pattern from a watermarked image is matched against all keys in the pool using $L_1$ distance in the embedding space, and the message associated with the closest key is selected as the predicted output.

\begin{figure}[t]
\centering
\includegraphics[width=0.9\linewidth]{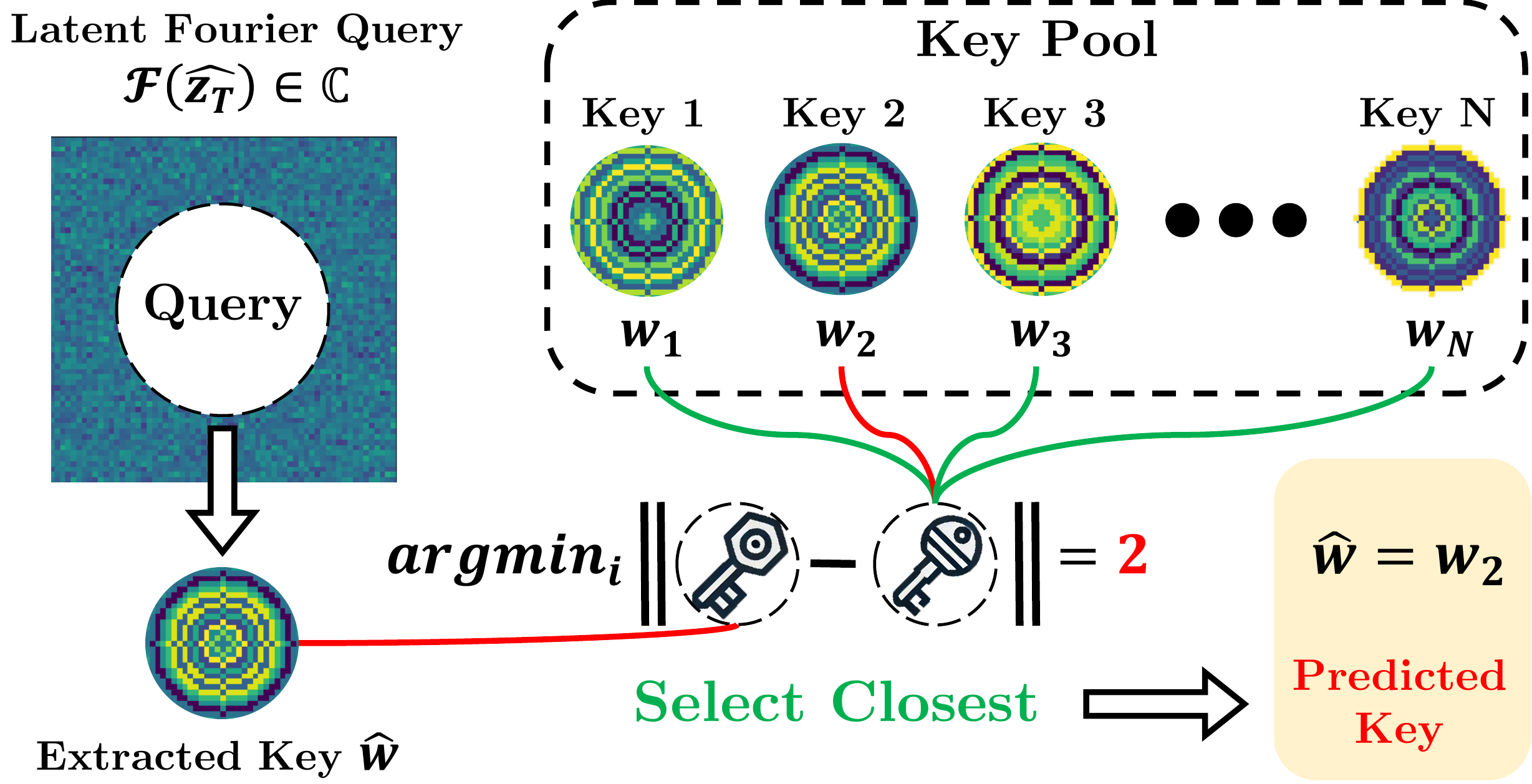}
\caption{Schematic illustration of the adapted identification protocol for Tree-Ring. Multiple candidate keys are generated and the extracted watermark is matched to the closest key based on $L_1$ distance.}
\label{fig:supple-TR-identify}
\end{figure}

\cref{fig:supple-TR-identify} provides a schematic overview of this process, illustrating how multiple keys are generated and compared in the identification pipeline. This procedure enables Tree-Ring to be evaluated under the same capacity-controlled setting as other semantic methods.
This protocol is also applied to other Gaussian radius-based methods, including Zodiac and the proposed HSTR.

\section{Additional Experimental Evidence}
\label{sec_supp:evidence}
\subsection{Processing Time and Detection Performance}
This section presents the processing time and detection performance (verification and identification) of different watermarking methods. As shown in \cref{tab:proc-time}, the \textit{merged-in-generation} approach does not introduce additional processing time since the watermarking process is inherently integrated into the diffusion-based image generation. The proposed method achieves superior detection performance without requiring additional processing time, demonstrating its efficiency.
On the other hand, \textit{post-hoc-based} approaches require per-image processing time, with some methods incurring significant computational costs. 
In particular, Zodiac demands several minutes per image as it requires multiple rounds of diffusion-based generation and latent vector optimization iterations for embedding the semantic watermark pattern, resulting in excessive computational overhead.

\begin{table}[t]
\centering
\footnotesize
\caption{Evaluation of watermarking methods based on processing time, verification, and identification performance. The best performance for each item is highlighted with shading, while bold text specifically marks the excessive processing time in Zodiac. \textit{Vrf.} and \textit{Idf.} denote the average detection performance in verification and identification tasks, respectively.}
\begin{tabular}{l|l|c|cc}
\toprule
\multicolumn{2}{c|}{Methods} & Processing Time $\downarrow$ & \textit{Vrf.} & \textit{Idf.}\\
\midrule
\multirow{4}{*}{\makecell{Post-hoc\\Based}} 
& DwtDct & 0.03 (s/img) & 0.637 & 0.083 \\
& DwtDctSvd & 0.07 (s/img) & 0.742 & 0.258 \\
& RivaGAN & 0.41 (s/img) & 0.857 & 0.482 \\
& Zodiac & \textbf{7.36 (m/img)} & 0.962 & 0.000 \\
\midrule
\multirow{5}{*}{\makecell{\\Merged in\\Generation}} 
& S.Sign. & 0.00 (s/img) & 0.836 & 0.265 \\
& Tree-Ring & 0.00 (s/img) & 0.655 & 0.114 \\
& RingID & 0.00 (s/img) & \cellcolor{gray!20}{0.997} & 0.964 \\
\cmidrule{2-5}
& HSTR (ours) & 0.00 (s/img) & 0.971 & 0.889\\
& HSQR (ours) & 0.00 (s/img) & \cellcolor{gray!20}{0.997} & \cellcolor{gray!20}{0.985} \\
\bottomrule
\end{tabular}
\label{tab:proc-time}
\end{table}

\begin{table}[th]
\centering
\scriptsize
\setlength{\tabcolsep}{4pt}
\caption{Verification and identification performance of Gaussian Shading under the same attack settings as in \cref{tab:verify-t@1f-ba} and \cref{tab:identify} of the main paper (MS-COCO only).}
\begin{tabular}{l|ccc|ccc}
\toprule
\multirow{2}{*}{\makecell{\\Attack Type}} 
& \multicolumn{3}{c|}{\textit{Verification}}
& \multicolumn{3}{c}{\textit{Identification}} \\
\cmidrule(rl){2-4}
\cmidrule(rl){5-7}
& G.Shading & HSTR & HSQR & G.Shading & HSTR & HSQR \\
\midrule
No Attack         & 1.000 & 1.000 & 1.000 & 1.000 & 1.000 & 1.000 \\
Bright.           & 0.962 & 0.899 & 0.991 & 0.522 & 0.714 & 0.958 \\
Cont.             & 1.000 & 1.000 & 1.000 & 0.998 & 0.999 & 1.000 \\
JPEG              & 0.992 & 0.994 & 1.000 & 0.724 & 0.886 & 0.994 \\
Blur              & 1.000 & 1.000 & 1.000 & 0.999 & 0.998 & 1.000 \\
Noise             & 0.997 & 0.806 & 0.983 & 0.919 & 0.460 & 0.901 \\
BM3D              & 0.999 & 0.999 & 1.000 & 0.926 & 0.972 & 0.999 \\
VAE-B             & 0.982 & 0.973 & 0.992 & 0.636 & 0.833 & 0.980 \\
VAE-C             & 0.987 & 0.982 & 1.000 & 0.657 & 0.831 & 0.987 \\
Diff.             & 0.999 & 0.997 & 1.000 & 0.827 & 0.971 & 0.999 \\
C.C.              & 0.998 & 1.000 & 1.000 & 0.658 & 1.000 & 1.000 \\
R.C.              & 1.000 & 1.000 & 1.000 & 0.986 & 1.000 & 1.000 \\
\midrule
Avg               & 0.993 & 0.971 & 0.997 & 0.821 & 0.889 & 0.985 \\
\bottomrule
\end{tabular}
\label{tab:supple-G.Shading}
\end{table}

\begin{table}[th]
\centering
\scriptsize
\caption{Normality assessment of latent distributions (1,000 samples). HSTR better preserves Gaussianity than Tree-Ring, as shown by standard deviation, KS p-value, and failure rate.}
\begin{tabular}{l|r@{\hspace{2mm}}c|c@{\hspace{3mm}}c}
\toprule
Methods & \multicolumn{1}{c}{Mean} & Std. Dev. & KS p-value $\uparrow$ & KS failure rate $\downarrow$ \\
\midrule
Tree-Ring & 0.0004 & 0.9620 & 0.2404 & 0.234 \\
HSTR (ours) & -0.0003 & 1.0000 & 0.4227 & 0.071 \\
\bottomrule
\end{tabular}
\label{tab:supple-gaussianity}
\end{table}

\subsection{Performance of Gaussian Shading as a Baseline}
\label{sec_supp:G.shading}
Following feedback received during review, this section presents the performance of Gaussian Shading (G.Shading) on the MS-COCO dataset. \cref{tab:supple-G.Shading} shows the detection results under the same attack settings as those used in \cref{tab:verify-t@1f-ba} and \cref{tab:identify} of the main paper. Since G.Shading is evaluated as a bitstream-based setting, we report Bit Accuracy for verification and Perfect Match Rate for identification as the detection metrics. In addition to detection performance, we also evaluate the generative quality as a supplement to \cref{tab:gen_quality}. G.Shading achieves an FID of 24.778 and a CLIP score of 0.330, which are comparable to those of our method HSQR (FID: 24.895, CLIP: 0.330).

\subsection{Preservation of Gaussianity}
We assess the normality of 1,000 latent samples, as reported in \cref{tab:supple-gaussianity}. Compared to Tree-Ring, HSTR more closely aligns with $\mathcal{N}(0,1)$ in terms of standard deviation and the Kolmogorov–Smirnov (KS) test. This includes higher p-values and lower failure rates, indicating stronger statistical consistency.

\begin{table}[t]
\centering
\scriptsize
\caption{Ablation study on the impact of Hermitian SFW and center-aware embedding in terms of robustness and generative quality.}
\begin{tabular}{@{\hspace{1mm}}c|c@{\hspace{2mm}}c|c@{\hspace{2mm}}c@{\hspace{2mm}}c@{\hspace{2mm}}c|c@{\hspace{3mm}}c@{\hspace{1mm}}}
\toprule
Case & SFW & Center & \textit{Signal.} & \textit{Regen.} & \textit{Crop.} & Avg & FID $\downarrow$ & CLIP $\uparrow$ \\
\midrule
A & \(\times\) & \(\times\) & 0.136 & 0.070 & 0.021 & 0.114 & 26.418 & 0.326 \\
B & \(\checkmark\) & \(\times\) & 0.856 & 0.812 & 0.374 & 0.777 & 25.071 & 0.329 \\
C & \(\checkmark\) & \(\checkmark\) & 0.838 & 0.878 & 1.000 & 0.889 & 25.062 & 0.329\\
\bottomrule
\end{tabular}
\label{tab:supple-disentangle}
\end{table}

\begin{table}[t]
\centering
\caption{Average PSNR, SSIM, and LPIPS values for each of the 11 attack types, computed over 1,000 MS-COCO generated images. These values reflect the typical level of distortion introduced by each attack.}
\begin{tabular}{l|ccc}
\toprule
Attack Type & PSNR $\uparrow$ & SSIM $\uparrow$ & LPIPS $\downarrow$ \\
\midrule
Bright.           & 28.421 & 0.558 & 0.383 \\
Cont.             & 28.015 & 0.824 & 0.092 \\
JPEG              & 32.909 & 0.898 & 0.066 \\
Blur              & 34.419 & 0.902 & 0.023 \\
Noise             & 44.926 & 0.894 & 0.113 \\
BM3D              & 35.648 & 0.910 & 0.074 \\
\midrule
VAE-B             & 33.594 & 0.884 & 0.093 \\
VAE-C             & 33.950 & 0.896 & 0.083 \\
Diff.             & 31.224 & 0.795 & 0.109 \\
\midrule
C.C.              & 30.959 & 0.503 & 0.431 \\
R.C.              & 33.164 & 0.702 & 0.298 \\
\bottomrule
\end{tabular}
\label{tab:supple-post-attack}
\end{table}

\subsection{Disentangling the Contributions}
We conduct an ablation analysis to examine the individual contributions of each component, with the results presented in \cref{tab:supple-disentangle}. From Tree-Ring (A), applying SFW (B) improves frequency integrity, which enhances robustness to signal and regeneration attacks as well as generative quality. Adding center-aware embedding (C), which corresponds to the proposed HSTR method, significantly enhances robustness to cropping attacks. Note that \textit{Signal.}, \textit{Regen.}, and \textit{Crop.} denote the average identification accuracy across the respective attack types. These results suggest that both components are necessary and complementary.

\subsection{Post-Attack Image Quality Assessment}
To complement the robustness evaluation, we provide an assessment of image quality degradation caused by various attacks.
\cref{tab:supple-post-attack} reports the average PSNR, SSIM, and LPIPS values computed over 1,000 MS-COCO generated images after applying each of the 11 attack types used in the main paper. We report PSNR and SSIM to measure pixel-level and structural similarity respectively, and include LPIPS to capture perceptual quality more closely aligned with human judgment. These metrics help ensure that attack strengths remain realistic and consistent across evaluation scenarios.

\begin{figure}[t]
\centering
\includegraphics[width=\linewidth]{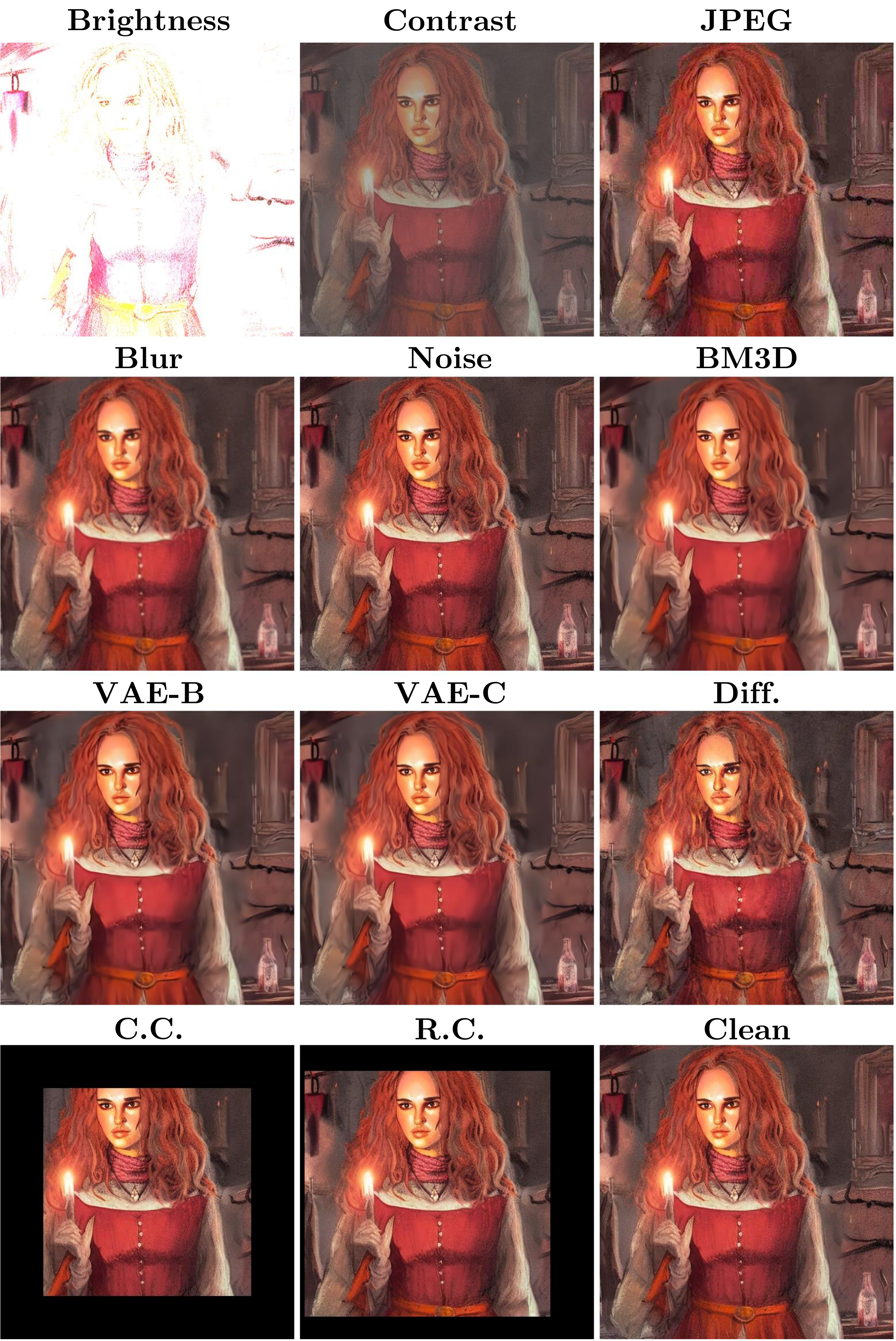}
\caption{Visual examples of all 11 attacks applied to a single clean image. The figure illustrates the perceptual effects of each attack type relative to the original input.}
\label{fig:supple-post-attack-Gustavo}
\end{figure}

In addition, \cref{fig:supple-post-attack-Gustavo} visualizes the effect of all 11 attacks on a single clean image, illustrating the diverse perceptual degradation introduced by each attack. Notably, the three examples in the third row of the figure, corresponding to regeneration attacks, appear visually high-quality from a classical signal processing perspective.
Despite the minimal perceptual degradation, many baseline methods fail to maintain correct detection under these attacks, as shown in \cref{tab:verify-t@1f-ba} (verification performance). This suggests that the attack strength is not weak, even if visual quality remains high. It also highlights that improving robustness to regeneration attacks remains a critical challenge, both for our method and for future research in this area.

\begin{table}[t]
\centering
\footnotesize
\caption{Verification and identification performance of semantic watermarking methods under diffusion-based regeneration attacks with varying noise steps. The results are based on 1,000 images generated from the MS-COCO dataset. Larger noise steps indicate stronger attack strength.}
\begin{tabular}{cl|ccccc}
\arrayrulecolor{black}
\toprule
\multirow{2}{*}{\makecell{\\Task}} & \multirow{2}{*}{\makecell{\\Methods}} &
\multicolumn{5}{c}{Noise Step} \\
\cmidrule(){3-7}
& & 20 & 60 & 100 & 140 & 180 \\
\midrule
\multirow{4}{*}{\textit{Vrf.}} 
& Tree-Ring & 0.701 & 0.543 & 0.404 & 0.317 & 0.262 \\
& HSTR (ours) & 1.000 & 0.997 & 0.998 & 0.987 & 0.980 \\
\cmidrule(l){2-7}
& RingID & 1.000 & 1.000 & 1.000 & 1.000 & 1.000 \\
& HSQR (ours) & 1.000 & 1.000 & 1.000 & 1.000 & 1.000 \\
\midrule
\multirow{4}{*}{\textit{Idf.}} 
& Tree-Ring & 0.144 & 0.054 & 0.034 & 0.014 & 0.012 \\
& HSTR (ours) & 0.999 & 0.971 & 0.926 & 0.868 & 0.781 \\
\cmidrule(l){2-7}
& RingID & 1.000 & 0.998 & 0.995 & 0.993 & 0.990 \\
& HSQR (ours) & 0.999 & 0.999 & 0.999 & 0.998 & 0.997 \\
\bottomrule
\end{tabular}
\label{tab:supple-diffattack-level}
\end{table}

\subsection{Ablation on Regeneration Strength}
To further investigate the robustness of our method against regeneration attacks, we conduct an ablation study by varying the noise strength in the diffusion-based regeneration attack. 
This experiment addresses the concern that the regeneration attacks used in the main paper may have been too weak compared to pixel-level Gaussian noise attacks, a typical signal processing perturbation.

Following the setup of Zhao et al.~\cite{zhao2025invisible}, the attack applies additive noise in the latent space using a formulation similar to the forward process in DDPM~\cite{ho2020denoising}:
\[
z_{t^*} \leftarrow \sqrt{\alpha(t^*)}z_0 + \sqrt{1 - \alpha(t^*)} \epsilon
\]
Here, \(z_0\) denotes the encoded latent representation of the image, and \(\epsilon\) is standard Gaussian noise. The variable \(t^*\) indicates the noise step that controls attack strength. The attacked image is then regenerated through the denoising process.

\cref{tab:supple-diffattack-level} reports the verification and identification performance of semantic watermarking methods that follow the \textit{merged-in-generation} scheme, evaluated under varying attack steps \(t^* \in \{20, 60, 100, 140, 180\}\). Following the main paper, verification is measured by TPR@1\%FPR and identification by accuracy, reported as Perfect Match Rate.
Detection performance progressively decreases as the noise step increases, indicating a corresponding increase in attack strength. A higher noise step results in lower detection performance, suggesting a stronger perturbation effect.
The proposed HSQR achieves the most robust detection performance under these stronger attack levels.
Focusing on Gaussian radius-based methods (Tree-Ring and HSTR), we observe that the proposed HSTR demonstrates significantly improved robustness over Tree-Ring.
This highlights the effectiveness of the Hermitian SFW component under challenging regeneration scenarios.

\section{Supplementary Experimental Results}
\label{sec_supp:supplements}
\begin{table*}[t]
\centering
\scriptsize 
\caption{Unified detection performance reported in terms of Bit Accuracy for all methods, including re-evaluation of semantic watermarks.}
\begin{tabular}{@{}c|l|cccccccccccc|c}
\arrayrulecolor{black}
\toprule
\multirow{2}{*}{\makecell{\\Datasets}} & \multirow{2}{*}{\makecell{\\Methods}} & No Attack &
\multicolumn{6}{c}{Signal Processing Attack} & 
\multicolumn{3}{c}{Regeneration Attack} & 
\multicolumn{2}{c|}{Cropping Attack} &
\multirow{2}{*}{\makecell{\\Avg}} \\
\cmidrule(rl){3-3}
\cmidrule(rl){4-9}
\cmidrule(rl){10-12}
\cmidrule(rl){13-14}
& & Clean & Bright. & Cont. & JPEG & Blur & Noise & BM3D & VAE-B & VAE-C & Diff. & C.C. & R.C. & \\
\midrule
\multirow{9}{*}{\makecell{\\MS-COCO}} 
& DwtDct & 0.863 & 0.572 & 0.522 & 0.516 & 0.677 & 0.859 & 0.532 & 0.523 & 0.521 & 0.519 & 0.729 & 0.810 & 0.637 \\
& DwtDctSvd & 1.000 & 0.555 & 0.473 & 0.602 & 1.000 & 1.000 & 0.784 & 0.648 & 0.596 & 0.644 & 0.744 & 0.861 & 0.742 \\
& RivaGAN & 0.999 & 0.862 & 0.986 & 0.821 & 0.998 & 0.969 & 0.934 & 0.570 & 0.552 & 0.608 & 0.991 & 0.995 & 0.857 \\
& S.Sign. & 0.995 & 0.894 & 0.978 & 0.806 & 0.911 & 0.721 & 0.838 & 0.717 & 0.715 & 0.478 & 0.987 & 0.991 & 0.836 \\
\cmidrule{2-15}
& Tree-Ring & 0.303 & 0.087 & 0.207 & 0.072 & 0.256 & 0.030 & 0.162 & 0.083 & 0.072 & 0.054 & 0.009 & 0.033 & 0.114 \\
& Zodiac & 0.000 & 0.000 & 0.000 & 0.000 & 0.000 & 0.000 & 0.000 & 0.000 & 0.000 & 0.000 & 0.000 & 0.000 & 0.000 \\
& HSTR (ours) & 1.000 & 0.714 & 0.999 & 0.886 & 0.998 & 0.460 & 0.972 & 0.833 & 0.831 & 0.971 & 1.000 & 1.000 & 0.889 \\
\cmidrule{2-15}
& RingID & 1.000 & 0.875 & 1.000 & 0.975 & 1.000 & 0.919 & 0.996 & 0.978 & 0.970 & 0.998 & 0.874 & 0.978 & 0.964 \\
& HSQR (ours) & 1.000 & 0.958 & 1.000 & 0.994 & 1.000 & 0.901 & 0.999 & 0.980 & 0.987 & 0.999 & 1.000 & 1.000 & 0.985 \\

\midrule
\multirow{9}{*}{\makecell{\\SD-Prompts}} 
& DwtDct & 0.819 & 0.557 & 0.516 & 0.506 & 0.685 & 0.822 & 0.530 & 0.513 & 0.512 & 0.509 & 0.723 & 0.794 & 0.624 \\
& DwtDctSvd & 1.000 & 0.537 & 0.459 & 0.610 & 0.999 & 0.998 & 0.859 & 0.659 & 0.620 & 0.623 & 0.743 & 0.860 & 0.747 \\
& RivaGAN & 0.991 & 0.823 & 0.963 & 0.810 & 0.988 & 0.961 & 0.915 & 0.572 & 0.535 & 0.567 & 0.980 & 0.983 & 0.841 \\
& S.Sign. & 0.994 & 0.899 & 0.967 & 0.769 & 0.888 & 0.742 & 0.809 & 0.677 & 0.671 & 0.493 & 0.983 & 0.990 & 0.824 \\
\cmidrule{2-15}
& Tree-Ring & 0.288 & 0.094 & 0.189 & 0.051 & 0.235 & 0.034 & 0.159 & 0.079 & 0.076 & 0.056 & 0.012 & 0.041 & 0.110 \\
& Zodiac & 0.000 & 0.000 & 0.000 & 0.000 & 0.000 & 0.000 & 0.000 & 0.000 & 0.000 & 0.000 & 0.000 & 0.000 & 0.000 \\
& HSTR (ours) & 1.000 & 0.655 & 0.999 & 0.863 & 0.999 & 0.555 & 0.980 & 0.846 & 0.847 & 0.973 & 1.000 & 1.000 & 0.893 \\
\cmidrule{2-15}
& RingID & 1.000 & 0.885 & 1.000 & 0.976 & 0.998 & 0.886 & 0.993 & 0.980 & 0.973 & 0.995 & 0.876 & 0.981 & 0.962 \\
& HSQR (ours) & 1.000 & 0.930 & 1.000 & 0.994 & 1.000 & 0.942 & 0.999 & 0.991 & 0.997 & 1.000 & 1.000 & 1.000 & 0.988 \\

\midrule
\multirow{9}{*}{\makecell{\\DiffusionDB}} 
& DwtDct & 0.842 & 0.563 & 0.515 & 0.509 & 0.672 & 0.829 & 0.526 & 0.513 & 0.514 & 0.512 & 0.723 & 0.801 & 0.627 \\
& DwtDctSvd & 0.998 & 0.558 & 0.463 & 0.593 & 0.997 & 0.995 & 0.830 & 0.658 & 0.608 & 0.621 & 0.742 & 0.860 & 0.744 \\
& RivaGAN & 0.987 & 0.839 & 0.960 & 0.790 & 0.985 & 0.937 & 0.893 & 0.553 & 0.518 & 0.556 & 0.974 & 0.979 & 0.831 \\
& S.Sign. & 0.990 & 0.890 & 0.967 & 0.787 & 0.889 & 0.726 & 0.819 & 0.690 & 0.687 & 0.496 & 0.981 & 0.986 & 0.826 \\
\cmidrule{2-15}
& Tree-Ring & 0.280 & 0.095 & 0.190 & 0.059 & 0.233 & 0.037 & 0.145 & 0.081 & 0.072 & 0.050 & 0.013 & 0.039 & 0.108 \\
& Zodiac & 0.000 & 0.000 & 0.000 & 0.000 & 0.000 & 0.000 & 0.000 & 0.000 & 0.000 & 0.000 & 0.000 & 0.000 & 0.000 \\
& HSTR (ours) & 0.996 & 0.721 & 0.992 & 0.854 & 0.989 & 0.563 & 0.958 & 0.830 & 0.821 & 0.952 & 0.996 & 0.996 & 0.889 \\
\cmidrule{2-15}
& RingID & 1.000 & 0.895 & 1.000 & 0.947 & 0.996 & 0.871 & 0.992 & 0.968 & 0.958 & 0.990 & 0.875 & 0.984 & 0.956 \\
& HSQR (ours) & 1.000 & 0.954 & 1.000 & 0.988 & 1.000 & 0.906 & 0.998 & 0.982 & 0.991 & 0.994 & 1.000 & 1.000 & 0.984 \\

\bottomrule
\end{tabular}
\label{tab:supple-ReportingBitAccuracy}
\end{table*}

\subsection{Reporting Bit Accuracy Results}
In the main paper, we use Bit Accuracy for verification and Perfect Match Rate for identification to evaluate the detection performance of bitstream-based approaches in \cref{sec:exp-baselines}. Following feedback received during review, we acknowledge that Bit Accuracy offers a more fine-grained perspective, particularly in high-capacity or multi-user settings. To complement the original results, we report unified detection performance in terms of Bit Accuracy across all methods in \cref{tab:supple-ReportingBitAccuracy}. Here, we adopt a strict evaluation criterion for the semantic methods: if the predicted pattern does not exactly match the ground-truth pattern, the Bit Accuracy for that sample is set to zero. We believe these results provide a more comprehensive view of overall detection performance.

\begin{table*}[t]
\centering
\scriptsize 
\caption{AUC values of semantic methods for verification, evaluated across different datasets and attack scenarios.}
\begin{tabular}{@{}c|l|cccccccccccc|c}
\arrayrulecolor{black}
\toprule
\multirow{2}{*}{\makecell{\\Datasets}} & \multirow{2}{*}{\makecell{\\Methods}} & No Attack &
\multicolumn{6}{c}{Signal Processing Attack} & 
\multicolumn{3}{c}{Regeneration Attack} & 
\multicolumn{2}{c|}{Cropping Attack} &
\multirow{2}{*}{\makecell{\\Avg}} \\
\cmidrule(rl){3-3}
\cmidrule(rl){4-9}
\cmidrule(rl){10-12}
\cmidrule(rl){13-14}
& & Clean & Bright. & Cont. & JPEG & Blur & Noise & BM3D & VAE-B & VAE-C & Diff. & C.C. & R.C. & \\
\midrule
\multirow{5}{*}{\makecell{\\MS-COCO}} 
& Tree-Ring & 0.997 & 0.895 & 0.990 & 0.923 & 0.994 & 0.870 & 0.977 & 0.912 & 0.924 & 0.921 & 0.913 & 0.962 & 0.940 \\
& Zodiac & 1.000 & 0.978 & 1.000 & 0.998 & 1.000 & 0.978 & 1.000 & 0.989 & 0.996 & 0.997 & 0.999 & 1.000 & 0.995 \\
& HSTR (ours) & 1.000 & 0.992 & 1.000 & 1.000 & 1.000 & 0.986 & 1.000 & 0.995 & 0.999 & 1.000 & 1.000 & 1.000 & 0.998 \\
\cmidrule{2-15}
& RingID & 1.000 & 0.999 & 1.000 & 1.000 & 1.000 & 0.998 & 1.000 & 0.994 & 1.000 & 1.000 & 1.000 & 1.000 & 0.999 \\
& HSQR (ours) & 1.000 & 0.999 & 1.000 & 1.000 & 1.000 & 0.996 & 1.000 & 0.997 & 1.000 & 1.000 & 1.000 & 1.000 & 0.999 \\

\midrule
\multirow{5}{*}{\makecell{\\SD-Prompts}} 
& Tree-Ring & 0.995 & 0.892 & 0.985 & 0.911 & 0.991 & 0.875 & 0.980 & 0.915 & 0.928 & 0.911 & 0.906 & 0.957 & 0.937 \\
& Zodiac & 1.000 & 0.940 & 1.000 & 0.998 & 1.000 & 0.977 & 1.000 & 0.985 & 0.998 & 0.995 & 1.000 & 1.000 & 0.991 \\
& HSTR (ours) & 1.000 & 0.979 & 1.000 & 1.000 & 1.000 & 0.986 & 1.000 & 0.996 & 0.999 & 1.000 & 1.000 & 1.000 & 0.997 \\
\cmidrule{2-15}
& RingID & 1.000 & 0.997 & 1.000 & 1.000 & 1.000 & 0.999 & 1.000 & 0.998 & 1.000 & 1.000 & 1.000 & 1.000 & 1.000 \\
& HSQR (ours) & 1.000 & 0.997 & 1.000 & 1.000 & 1.000 & 0.997 & 1.000 & 0.999 & 1.000 & 1.000 & 1.000 & 1.000 & 0.999 \\

\midrule
\multirow{5}{*}{\makecell{\\DiffusionDB}} 
& Tree-Ring & 0.993 & 0.894 & 0.983 & 0.902 & 0.988 & 0.856 & 0.971 & 0.905 & 0.912 & 0.900 & 0.904 & 0.955 & 0.930 \\
& Zodiac & 0.997 & 0.958 & 0.997 & 0.990 & 0.997 & 0.942 & 0.996 & 0.975 & 0.984 & 0.983 & 0.993 & 0.997 & 0.984 \\
& HSTR (ours) & 1.000 & 0.988 & 1.000 & 0.999 & 1.000 & 0.974 & 1.000 & 0.995 & 0.998 & 1.000 & 1.000 & 1.000 & 0.996 \\
\cmidrule{2-15}
& RingID & 1.000 & 0.999 & 1.000 & 1.000 & 1.000 & 0.994 & 1.000 & 0.997 & 1.000 & 1.000 & 1.000 & 1.000 & 0.999 \\
& HSQR (ours) & 1.000 & 0.997 & 1.000 & 1.000 & 1.000 & 0.994 & 1.000 & 0.999 & 1.000 & 1.000 & 1.000 & 1.000 & 0.999 \\
\bottomrule
\end{tabular}
\label{tab:verify-auc}
\end{table*}

\begin{table*}[t]
\centering
\scriptsize 
\caption{Maximum verification accuracy for semantic methods across different datasets and attack scenarios.}
\begin{tabular}{@{}c|l|cccccccccccc|c}
\arrayrulecolor{black}
\toprule
\multirow{2}{*}{\makecell{\\Datasets}} & \multirow{2}{*}{\makecell{\\Methods}} & No Attack &
\multicolumn{6}{c}{Signal Processing Attack} & 
\multicolumn{3}{c}{Regeneration Attack} & 
\multicolumn{2}{c|}{Cropping Attack} &
\multirow{2}{*}{\makecell{\\Avg}} \\
\cmidrule(rl){3-3}
\cmidrule(rl){4-9}
\cmidrule(rl){10-12}
\cmidrule(rl){13-14}
& & Clean & Bright. & Cont. & JPEG & Blur & Noise & BM3D & VAE-B & VAE-C & Diff. & C.C. & R.C. & \\
\midrule
\multirow{5}{*}{\makecell{\\MS-COCO}}
& Tree-Ring & 0.979 & 0.828 & 0.959 & 0.852 & 0.968 & 0.808 & 0.930 & 0.846 & 0.856 & 0.849 & 0.850 & 0.911 & 0.886 \\
& Zodiac & 0.998 & 0.931 & 0.998 & 0.984 & 0.998 & 0.941 & 0.998 & 0.968 & 0.977 & 0.983 & 0.992 & 0.997 & 0.980 \\
& HSTR (ours) & 1.000 & 0.963 & 1.000 & 0.993 & 1.000 & 0.943 & 0.999 & 0.982 & 0.988 & 0.997 & 1.000 & 1.000 & 0.989 \\
\cmidrule{2-15}
& RingID & 1.000 & 0.991 & 1.000 & 1.000 & 1.000 & 0.990 & 1.000 & 0.996 & 1.000 & 1.000 & 1.000 & 1.000 & 0.998 \\
& HSQR (ours) & 1.000 & 0.992 & 1.000 & 1.000 & 1.000 & 0.988 & 1.000 & 0.996 & 1.000 & 1.000 & 1.000 & 1.000 & 0.998 \\

\midrule
\multirow{5}{*}{\makecell{\\SD-Prompts}} 
& Tree-Ring & 0.973 & 0.822 & 0.951 & 0.835 & 0.962 & 0.805 & 0.933 & 0.843 & 0.857 & 0.838 & 0.847 & 0.898 & 0.880 \\
& Zodiac & 0.998 & 0.888 & 0.999 & 0.986 & 0.999 & 0.954 & 0.998 & 0.967 & 0.983 & 0.978 & 0.994 & 0.998 & 0.978 \\
& HSTR (ours) & 1.000 & 0.939 & 1.000 & 0.992 & 1.000 & 0.945 & 0.998 & 0.989 & 0.989 & 0.998 & 1.000 & 1.000 & 0.987 \\
\cmidrule{2-15}
& RingID & 1.000 & 0.984 & 1.000 & 1.000 & 1.000 & 0.991 & 1.000 & 0.998 & 1.000 & 1.000 & 1.000 & 1.000 & 0.998 \\
& HSQR (ours) & 1.000 & 0.980 & 1.000 & 0.999 & 1.000 & 0.993 & 1.000 & 0.998 & 1.000 & 1.000 & 1.000 & 1.000 & 0.997 \\

\midrule
\multirow{5}{*}{\makecell{\\DiffusionDB}} 
& Tree-Ring & 0.968 & 0.823 & 0.944 & 0.833 & 0.952 & 0.795 & 0.919 & 0.839 & 0.844 & 0.827 & 0.843 & 0.897 & 0.874 \\
& Zodiac & 0.993 & 0.903 & 0.991 & 0.964 & 0.992 & 0.920 & 0.989 & 0.951 & 0.964 & 0.952 & 0.981 & 0.991 & 0.966 \\
& HSTR (ours) & 0.999 & 0.951 & 0.998 & 0.988 & 0.997 & 0.917 & 0.994 & 0.980 & 0.982 & 0.992 & 0.999 & 0.999 & 0.983 \\
\cmidrule{2-15}
& RingID & 1.000 & 0.991 & 1.000 & 0.999 & 1.000 & 0.978 & 1.000 & 0.998 & 1.000 & 1.000 & 1.000 & 1.000 & 0.997 \\
& HSQR (ours) & 1.000 & 0.987 & 1.000 & 0.999 & 1.000 & 0.983 & 1.000 & 0.997 & 0.999 & 1.000 & 1.000 & 1.000 & 0.997 \\
\bottomrule
\end{tabular}
\label{tab:verify-maxacc}
\end{table*}

\begin{figure*}[t]
\centering
\includegraphics[width=\linewidth, height=0.98\textheight]{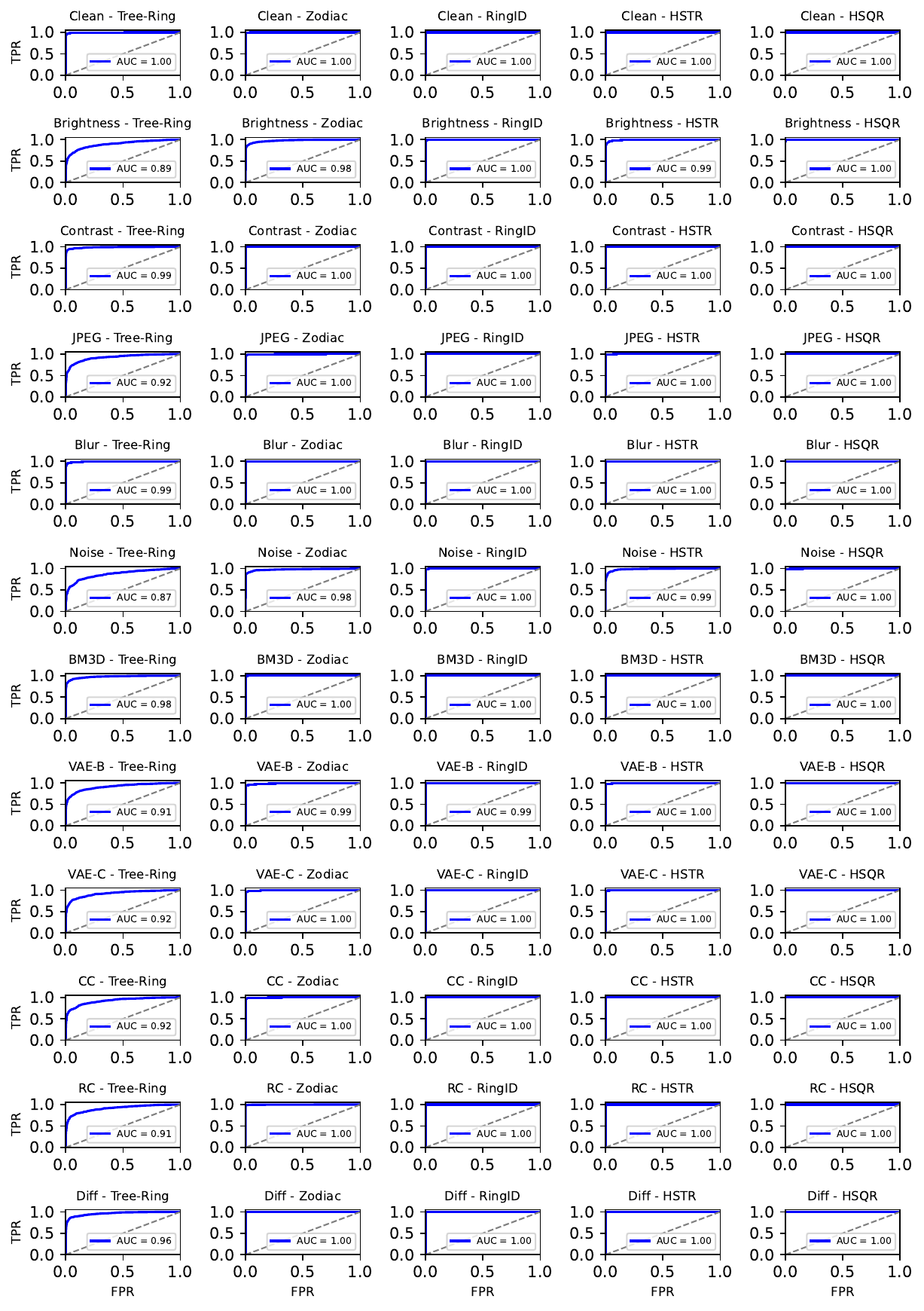}
\caption{ROC curve for verification performance on MS-COCO under different attack scenarios.}
\label{fig:roc-coco}
\end{figure*}

\begin{figure*}[t]
\centering
\includegraphics[width=\linewidth, height=0.98\textheight]{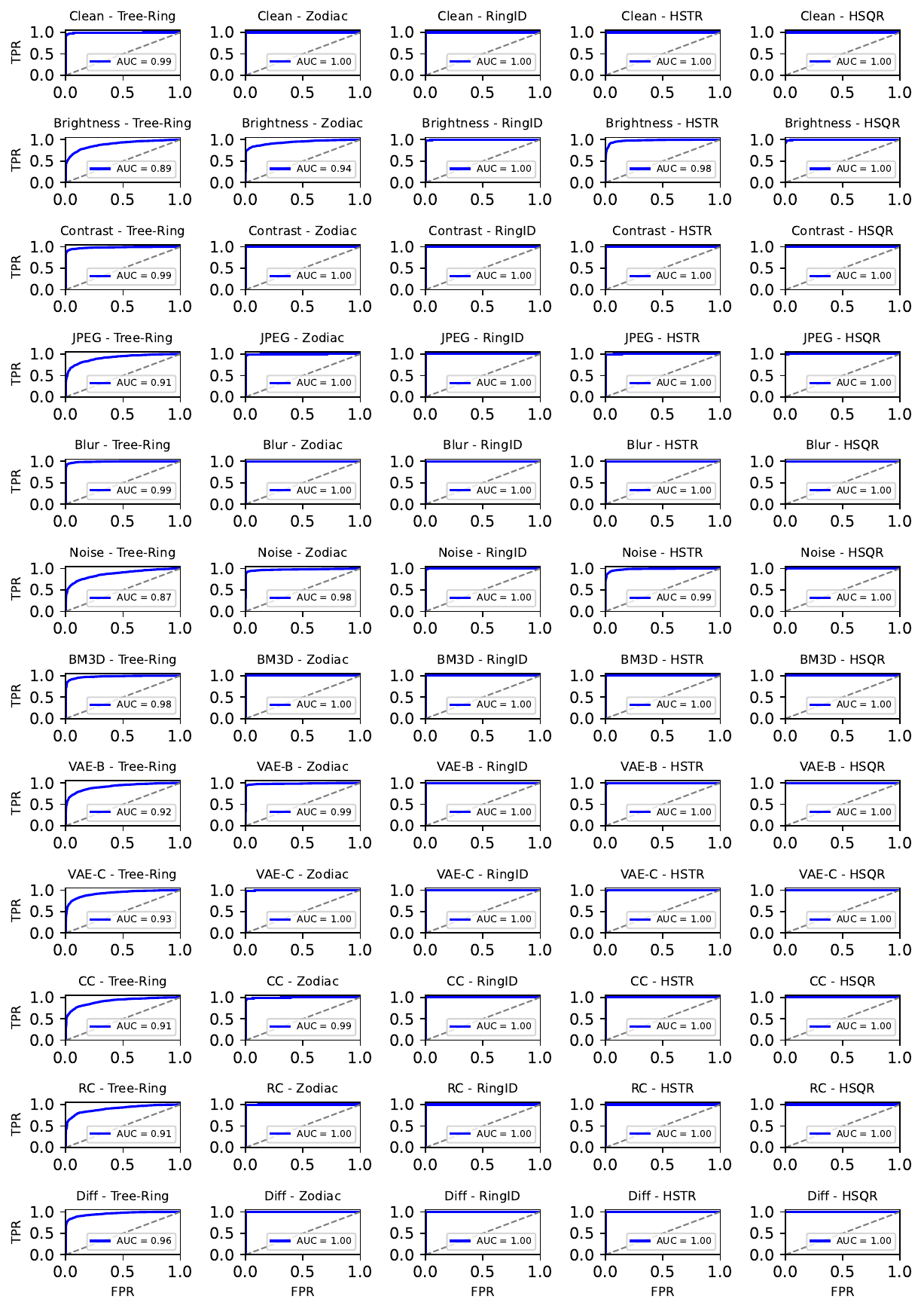}
\caption{ROC curve for verification performance on SD-Prompts under different attack scenarios.}
\label{fig:roc-Gustavo}
\end{figure*}

\begin{figure*}[t]
\centering
\includegraphics[width=\linewidth, height=0.98\textheight]{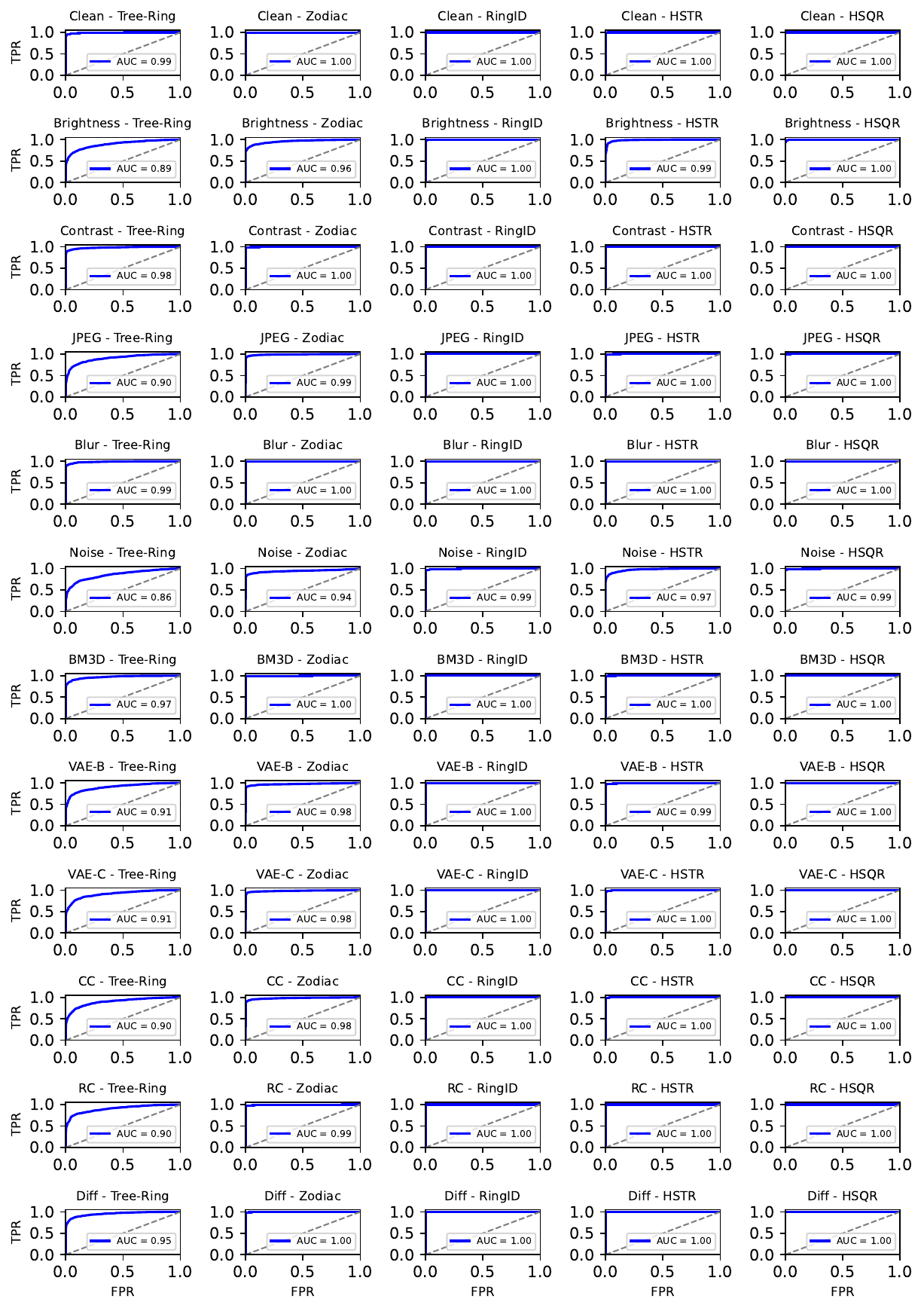}
\caption{ROC curve for verification performance on DiffusionDB under different attack scenarios.}
\label{fig:roc-DB1k}
\end{figure*}

\subsection{Further Results for Verification}
This section provides supplementary results for semantic methods on the verification task introduced in \cref{sec:exp-baselines}.
\begin{itemize}
    \item \cref{fig:roc-coco}, \cref{fig:roc-Gustavo}, and \cref{fig:roc-DB1k} illustrate the Receiver Operating Characteristic (ROC) curves for different datasets under various attack scenarios.
    \item \cref{tab:verify-auc} and \cref{tab:verify-maxacc} summarize the corresponding Area Under the Curve (AUC) values and maximum accuracy for each dataset.
\end{itemize}

\begin{table*}[t]
\centering
\footnotesize
\caption{Detailed verification and identification performance for Hermitian SFW ablation cases, supplementing \cref{tab:ablation-scope} in the main paper.}
\begin{tabular}{cc|cccccccccccc|c}
\arrayrulecolor{black}
\toprule
\multirow{2}{*}{\makecell{\\Task}} & \multirow{2}{*}{\makecell{\\Case}} & No Attack &
\multicolumn{6}{c}{Signal Processing Attack} & 
\multicolumn{3}{c}{Regeneration Attack} & 
\multicolumn{2}{c|}{Cropping Attack} &
\multirow{2}{*}{\makecell{\\Avg}} \\
\cmidrule(rl){3-3}
\cmidrule(rl){4-9}
\cmidrule(rl){10-12}
\cmidrule(rl){13-14}
& & Clean & Bright. & Cont. & JPEG & Blur & Noise & BM3D & VAE-B & VAE-C & Diff. & C.C. & R.C. & \\
\midrule
\multirow{4}{*}{\textit{Vrf.}} 
& A & 0.957 & 0.452 & 0.900 & 0.548 & 0.934 & 0.412 & 0.815 & 0.501 & 0.536 & 0.509 & 0.734 & 0.543 & 0.653 \\
& B & 1.000 & 0.601 & 1.000 & 0.772 & 1.000 & 0.588 & 0.977 & 0.737 & 0.774 & 0.550 & 0.853 & 0.810 & 0.805 \\
& C & 1.000 & 0.769 & 1.000 & 0.975 & 1.000 & 0.627 & 0.991 & 0.931 & 0.920 & 1.000 & 1.000 & 0.990 & 0.936 \\
& D & 1.000 & 0.899 & 1.000 & 0.994 & 1.000 & 0.806 & 0.999 & 0.981 & 0.982 & 1.000 & 1.000 & 0.997 & 0.971 \\
\midrule
\multirow{4}{*}{\textit{Idf.}} 
& A & 0.303 & 0.090 & 0.207 & 0.072 & 0.256 & 0.030 & 0.162 & 0.084 & 0.072 & 0.009 & 0.033 & 0.054 & 0.114 \\
& B & 0.982 & 0.320 & 0.854 & 0.268 & 0.904 & 0.106 & 0.624 & 0.306 & 0.293 & 0.018 & 0.109 & 0.202 & 0.416 \\
& C & 0.997 & 0.505 & 0.984 & 0.687 & 0.980 & 0.212 & 0.837 & 0.631 & 0.613 & 1.000 & 1.000 & 0.852 & 0.775 \\
& D & 1.000 & 0.714 & 0.999 & 0.886 & 0.998 & 0.460 & 0.972 & 0.841 & 0.831 & 1.000 & 1.000 & 0.971 & 0.889 \\
\bottomrule
\end{tabular}
\label{tab:supple-scope}
\end{table*}

\begin{table}[t]
\centering
\scriptsize 
\caption{Identification accuracy under center crop and random crop attacks at different crop scales. These results correspond to \cref{fig:ablation-crop}}
\begin{tabular}{l|ccccccc}
\arrayrulecolor{black}
\toprule
\multicolumn{8}{c}{\textbf{\textit{Center Crop Attack}}} \\
\midrule
\multirow{2}{*}{\makecell{\\Methods}} &
\multicolumn{6}{c}{Crop Scale} \\
\cmidrule(){2-8}
& 0.2 & 0.3 & 0.4 & 0.5 & 0.6 & 0.7 & 0.8 \\
\midrule
RingID & 0.153 & 0.369 & 0.647 & 0.874 & 0.934 & 0.974 & 0.992 \\
HSTR (ours) & 0.818 & 0.997 & 1.000 & 1.000 & 1.000 & 1.000 & 1.000 \\
HSQR (ours) & 0.555 & 0.998 & 1.000 & 1.000 & 1.000 & 1.000 & 1.000 \\
\bottomrule
\toprule
\multicolumn{8}{c}{\textbf{\textit{Random Crop Attack}}} \\
\midrule
\multirow{2}{*}{\makecell{\\Methods}} &
\multicolumn{6}{c}{Crop Scale} \\
\cmidrule(){2-8}
& 0.2 & 0.3 & 0.4 & 0.5 & 0.6 & 0.7 & 0.8 \\
\midrule
RingID & 0.496 & 0.559 & 0.774 & 0.919 & 0.971 & 0.970 & 0.997 \\
HSTR (ours) & 0.489 & 0.903 & 0.992 & 0.999 & 1.000 & 1.000 & 1.000 \\
HSQR (ours) & 0.955 & 0.999 & 0.999 & 1.000 & 1.000 & 1.000 & 1.000 \\
\bottomrule
\end{tabular}
\label{tab:supple-crop}
\end{table}

\begin{table}[t]
\centering
\scriptsize 
\caption{Average identification accuracy across watermark message capacities. The values are computed over all attack scenarios for semantic methods. These results correspond to \cref{fig:ablation-capacity}}
\begin{tabular}{l|cccccc}
\arrayrulecolor{black}
\toprule
\multirow{3}{*}{Methods}
& \multicolumn{6}{c}{Embedding Density ($10^{-5}$ bpp)} \\
\cmidrule(){2-7}
& 2.29 & 3.05 & 3.81 & 4.20 & 4.58 & 4.96 \\
\midrule
Tree-Ring & 0.338 & 0.271 & 0.136 & 0.114 & 0.083 & 0.064 \\
Zodiac & 0.027 & 0.000 & 0.000 & 0.000 & 0.000 & 0.000 \\
HSTR (ours) & 0.960 & 0.936 & 0.913 & 0.889 & 0.881 & 0.862 \\
\midrule
RingID & 0.995 & 0.989 & 0.978 & 0.964 & 0.940 & 0.888 \\
HSQR (ours) & 0.993 & 0.990 & 0.987 & 0.985 & 0.984 & 0.981 \\
\bottomrule
\end{tabular}
\label{tab:supple-capacity}
\end{table}

\subsection{Numerical Results for Ablation Study}
This section presents the numerical data corresponding to the figures in \cref{sec:ablations} (Ablation Study). 
\begin{itemize}
    \item \cref{tab:supple-scope} provides detailed results for the ablation study on Hermitian SFW cases presented in \cref{tab:ablation-scope}.
    \item \cref{tab:supple-crop} shows the identification accuracy under center crop and random crop attacks at different crop scales, corresponding to \cref{fig:ablation-crop}.
    \item \cref{tab:supple-capacity} presents the average identification accuracy across clean conditions and 11 attack scenarios for different watermarking capacities, as shown in \cref{fig:ablation-capacity}. For each capacity, we report its associated embedding density, computed based on a fixed image resolution of $512\times512$. This supplements the capacity-related analysis by making the notion of embedding density explicit, as suggested during the review.
\end{itemize}

\begin{figure*}[t]
\centering
\includegraphics[height=0.95\textheight]{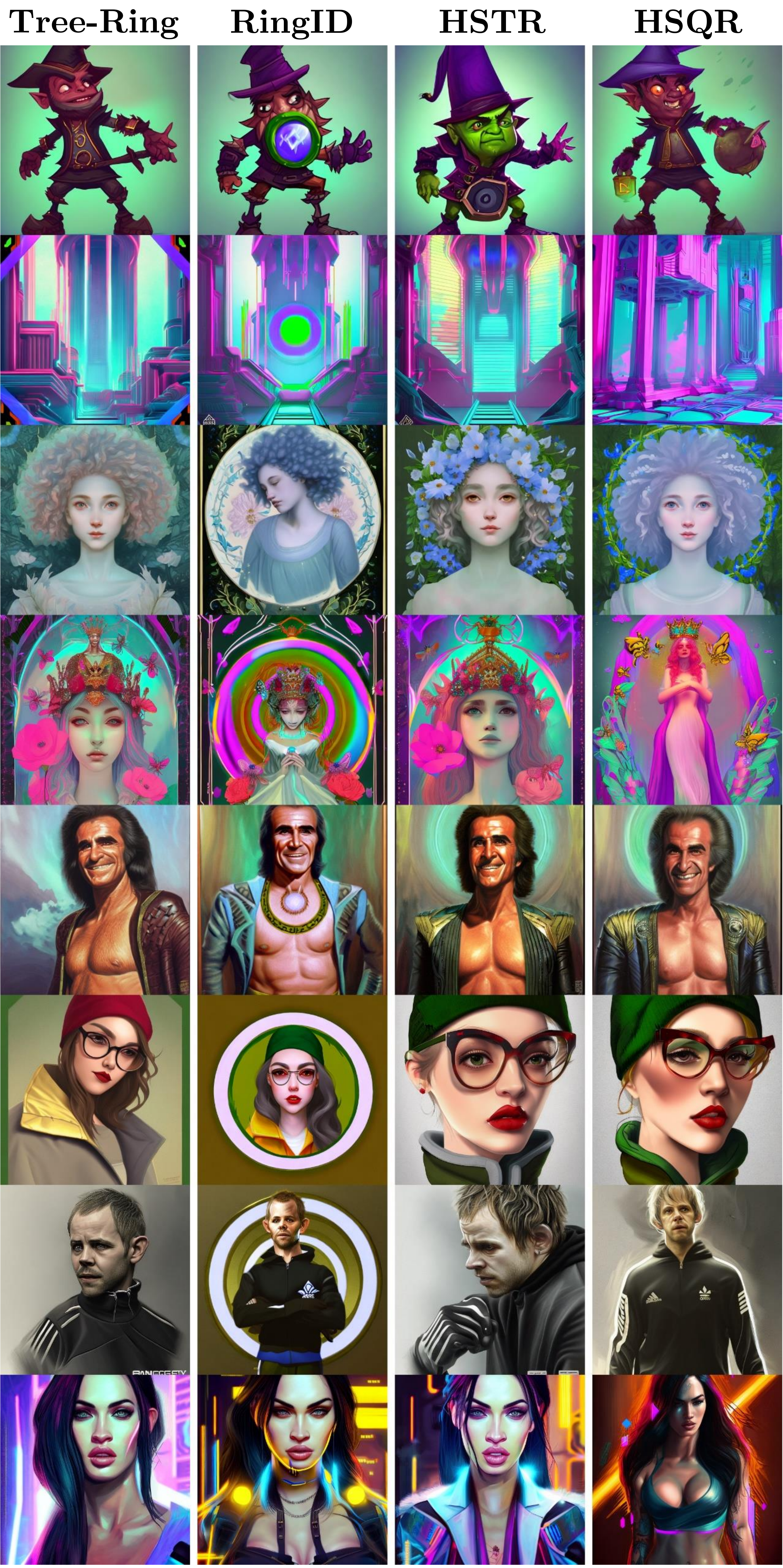}
\caption{Qualitative comparison of semantic watermarking methods following the \textit{merged-in-generation} scheme. The generated images are produced from the same prompt, illustrating the visual differences across different watermarking approaches.}
\label{fig:qualitative}
\end{figure*}

\subsection{Qualitative Analysis for Semantic Methods}
This section presents qualitative results for semantic watermarking methods following the \textit{merged-in-generation} scheme, showcasing generated images from the same prompt, as illustrated in \cref{fig:qualitative}.
In the case of RingID, excessive emphasis on detection performance at the expense of image quality results in imbalanced trade-offs, causing noticeable \textit{ring-like} artifacts in the generated images. This phenomenon aligns with the low CLIP score (0.324) reported in \cref{tab:gen_quality}, indicating degraded text-image alignment.
Furthermore, for Tree-Ring, which employs a Gaussian radius-based watermark pattern, the proposed method HSTR preserves Fourier integrity while embedding the same pattern. As a result, HSTR improves the CLIP score from 0.326 to 0.329, demonstrating its ability to enhance the quality of diffusion-generated images.

To further assess perceptual quality, we conduct a Mean Opinion Score (MOS) study based on human evaluations. For each of 10 prompts, we present the corresponding images generated by Tree-Ring, RingID, HSTR, and HSQR as a group, and ask 10 human evaluators to rate each image individually on a scale from 1 (very poor) to 5 (excellent). Participants assign a separate score to each image based on its visual quality.
The resulting average MOS scores are 2.82 for RingID, 3.54 for Tree-Ring, 3.69 for HSQR, and 3.86 for HSTR. While HSTR ranked highest in MOS, HSQR remains strong across both human ratings and CLIP/FID metrics, showing consistent perceptual quality overall.

\section{Outlook and Deployment Considerations}
\label{sec_supp:deployment}
The growing accessibility of LDMs has enabled an unprecedented scale of generative content creation. As synthetic media becomes ubiquitous, embedding provenance signals at generation time, rather than through costly post-processing, will become increasingly vital.

Meanwhile, modern NPUs are optimized for low-power, high-throughput AI inference. These accelerators favor low-precision formats such as FP16 or INT8, aligning well with the inference-only use of lightweight generative models like Stable Diffusion.

Our proposed watermarking methods, HSTR and HSQR, are inherently compatible with this direction. They require no additional training, integrate seamlessly into the generation pipeline, and avoid post-hoc overhead. This \textit{merged-in-generation} design, combined with semantic robustness and compatibility with quantized LDMs, positions our approach as a strong candidate for deployment in scalable, energy-efficient environments.